\documentclass{article}

\PassOptionsToPackage{numbers}{natbib}

\usepackage[preprint]{neurips_style}

\usepackage[utf8]{inputenc} 
\usepackage[T1]{fontenc}    
\usepackage{url}            
\usepackage{booktabs}       
\usepackage{amsfonts}       
\usepackage{nicefrac}       
\usepackage{microtype}      
\usepackage{xcolor}         

\usepackage{jcd}
\bluehyperref
\usepackage{algorithm}
\usepackage{algorithmic}
\usepackage{upgreek}
\usepackage{overpic}
\usepackage{pdfpages}

\usepackage{xr}
\newcommand{\score}{s}

\newcommand{\scorerv}{S}

\newcommand{\biject}{\mathfrak{S}}

\newcommand{\weakset}{W}

\newcommand{\quantile}{\mc{Q}}

\newcommand{\probe}{\varphi_i}

\usepackage{multirow}
\usepackage{tikz,forest}
\usetikzlibrary{arrows.meta}

\forestset{
    .style={
        for tree={
            base=bottom,
            child anchor=north,
            align=center,
            s sep+=1cm,
    straight edge/.style={
        edge path={\noexpand\path[\forestoption{edge},thick,-{Latex}] 
        (!u.parent anchor) -- (.child anchor);}
    },
    if n children={0}
        {tier=word, draw, thick, rectangle}
        {draw, diamond, thick, aspect=2},
    if n=1{%
        edge path={\noexpand\path[\forestoption{edge},thick,-{Latex}] 
        (!u.parent anchor) -| (.child anchor) node[pos=.2, above] {Y};}
        }{
        edge path={\noexpand\path[\forestoption{edge},thick,-{Latex}] 
        (!u.parent anchor) -| (.child anchor) node[pos=.2, above] {N};}
        }
        }
    }
}

\title{Query-Adaptive Predictive Inference with Partial Labels}

\author{%
  Maxime Cauchois \\
  Department of Statistics\\
  Stanford University\\
  Stanford, CA 94305 \\
  \texttt{maxcauch@stanford.edu} \\
   \And
   John Duchi \\
   Department of Statistics and Electrical Engineering \\
   Stanford University \\
   Stanford, CA 94305 \\
   \texttt{jduchi@stanford.edu} \\
}

\begin{document}

\maketitle

\begin{abstract}
The cost and scarcity of fully supervised labels in statistical machine
learning encourage using partially labeled data for model validation
as a cheaper and more accessible alternative. Effectively collecting and
leveraging weakly supervised data for large-space structured prediction
tasks thus becomes an important part of an end-to-end learning system. We propose a new computationally-friendly methodology to construct predictive
sets using only partially labeled data on top of  black-box
predictive models.  To do so, we introduce ``probe'' functions as a way to describe weakly supervised instances and define a false discovery
proportion-type loss, both of which seamlessly adapt to partial supervision
and structured prediction---ranking, matching, segmentation, multilabel or multiclass classification. Our experiments highlight the validity of our predictive set
construction as well as the attractiveness of a more flexible user-dependent loss framework.

\end{abstract}

\section{Introduction}
\label{sec:introduction}

Consider supervised learning problems, where one aims to predict a target $Y \in \mc{Y}$ using features $X \in \mc{X}$.  The traditional
pipeline is to collect fully supervised examples $\{X_j, Y_j \}_{j=1}^N$,
then train a predictive function on this data.  Yet this is only part of the
story.  As these models make increasingly critical decisions, e.g.\ in
healthcare, autonomous vehicles, or finance, we wish to release predictions
with actionable error guarantees and precise uncertainty quantification.
Typically, this involves an additional calibration step and requires
additional (expensive) fresh data~\cite{BarberCaRaTi19a}.

However, collecting fully supervised instances for calibration
is a strong ask.  First, they can be costly to acquire, so that it makes
more sense to only use them for training; additionally, they may simply be
inaccessible, as when labelers only provide partial feedback
about their own label $Y \in \mc{Y}$. We refer to a
putative labeler as a ``user,'' taking as motivation
several examples: a shopping
application where users reveal a small number of their preferences
(what they purchase)
instead of a ranking of all items for sale;
physicians may partially segment a scan between
healthy tissues and tumor to save time;
in multilabel image classification, labelers
may identify a subset of present objects.

Our goal is to leverage the partial feedback we can collect to construct
valid predictive sets containing the ``true'' label $Y$ with a prescribed
probability. Our focus is on structured prediction tasks with large output
space $\mc{Y}$---e.g.\ $\{-1,1\}^K$ in multilabel or segmentation
tasks, or the space of permutations over $K$ elements ($\biject_K$) in
ranking or matching predictions--- so that listing all relevant
configurations to give validity is impractical.  We thus refocus on
computationally efficient ways to describe label configurations of interest,
formalizing the idea via a (problem-dependent) index set $\mc{I}_\mc{Y}$
and collection $\{ \probe \}_{i \in \mc{I}_\mc{Y}} \subset \{-1,1\}^{\mc{Y}}$ of
\emph{probe} (or ``query'') functions satisfying the identifiability requirement
\begin{align}
  \label{eqn:probe-fcn-identifiable}
  \text{for any pair} ~ y, y' \in \mc{Y},
  ~ \mbox{if}~ \probe(y) = \probe(y')
  ~~ \mbox{for~all~} i \in \mc{I}_\mc{Y},
  \text{ then } y = y'.
\end{align}
There are often multiple choices of probes. In the ranking case, where $y
\in \mc{Y} = \biject_K$ is a permutation with $y(k)$ the $k$-th highest item, one can choose $\varphi_{i,j}(y) =
\sign\left(y^{-1}(j)-y^{-1}(i)\right)$ for $i<j$, which describes a
ranking through its binary comparisons; conversely, 
$\varphi_{i,j}(y) = 2 \cdot
\indic{y(i)=j}-1$ gives $\varphi_{ij}(1) = 1$ if item $j$ has rank $i$, $-1$ otherwise.
In hierarchical classification problems,
where each label is the leaf of a tree $\mc{T} = (V,E)$, each node $v
\in V$ then represents $\varphi_v(y) \defeq \indic{v \text{
    is an ancestor of } y}$, which characterizes the label $y$ via its path
from the root.  Examples such as segmentation and multilabel prediction
have label spaces of the form $y \in \{-1, 1\}^K$ (where
$y_i$ indicates the segmentation of pixel $i$ in the former
and whether object $i$ appears in the latter) admit
the straightforward probes $\varphi_i(y) = y_i$.

Assuming the existence of such a probe set, we propose a generic method to produce valid predictive sets. 
Examples consist of a triple $(x,
y, I) \in \mc{X} \times \mc{Y} \times 2^{\mc{I}_{\mc{Y}}}$, where the index set $I \subset \mc{I}_\mc{Y}$ specifies queries. 
Given a sample $(X_i, Y_i, I_i)_{i = 1}^n$,
our method returns a confidence set mapping $\what{C}_{n}$ that on
a new example $(X_{n+1}, Y_{n+1}, I_{n+1})$ answers
(potentially, a subset of) the queries in $I_{n+1}$ while guaranteeing
\begin{align}
\label{eqn:partial-fpp-loss-quantile-coverage}
\P \left[ \dfrac{\text{Number of correctly answered queries by}~\what{C}_n(X_{n+1})}{\text{Number of queries answered by}~\what{C}_n(X_{n+1})} \ge 1-\delta \right] \ge 1-\alpha.
\end{align}
Here, users only provide partial feedback through a random subset $\{
\probe(y) \}_{i \in I}$ of probe functions ($I \subset \mc{I}_\mc{Y}$), and
the method may only observe $(X, \{ \probe(Y) \}_{i \in I})$.

\citet{CauchoisGuAlDu22} initiate the study of predictive inference with
weak supervision,  summarizing partial labels of instances
as a (potentially very large) ``weak'' set $W \subset \mc{Y}$, which
here corresponds to
\begin{align*}
  \weakset \defeq \left\{ y' \in \mc{Y}  \mid
  \probe(y) = \probe(y')  ~ \mbox{for~all~} i \in I \right\} \subset \mc{Y}.
\end{align*}
Our methods provide guarantees for prediction in this weak
supervision regime.  Given a sample $\{(X_i, Y_i, I_i)\}_{i=1}^n$ of size $n
\ge 1$, \citet{CauchoisGuAlDu22} show that, in distinction
from conformal inference
methods in the fully supervised regime~\cite{CauchoisGuDu21, SadinleLeWa19,
  VovkGaSh05, RomanoSeCa20, RomanoPaCa19}, any predictive set function $\what{C}_n : \mc{X} \toto \mc{Y}$ providing distribution free
coverage while only using partially labeled samples
\emph{cannot} satisfy the standard fully supervised coverage guarantee
\begin{align}
\label{eqn:marginal-conformal-coverage-full}
\P\left( Y_{n+1} \in \what{C}_n(X_{n+1}) \right) \ge 1-\alpha,
\end{align}
unless the confidence set is so large that it carries no information.  They
resolve to an analogous notion of ``weak'' coverage, where any label that
belongs to $\weakset$ guarantees
\begin{align}
\label{eqn:marginal-conformal-coverage-weak-coverage}
\P\left( \what{C}_n(X_{n+1}) \cap W_{n+1} \neq \emptyset \right) \ge 1-\alpha.
\end{align}
While less conservative, this approach still has the primary drawback that
weak coverage effectively remains a $0$-$1$ loss: for every instance,
there is no middle ground between a predictive set's validity or
invalidity. This is undesirable in structured prediction tasks, like
ranking, segmentation, or matching, as it typically results in large
uninformative predictive sets.  Additionally, only requiring that $\weakset
\cap \what{C}_n(X) \neq \emptyset$ places a strong emphasis on the size of
the weak set itself, making it much easier to provide weak coverage for
instances with the least supervision (when $\weakset$ is large).

We take a two-pronged approach to tackle these issues, building off
of~\citeauthor{BatesAnLeMaJo21}'s risk-controlling predictive sets for fully
supervised tasks~\cite{BatesAnLeMaJo21}.  
We replace the 0-1 loss $\ell_{01}(C) \defeq \indic{C
  \cap W \neq 0}$ with a more general loss $\ell : 2^\mc{Y} \to [0,1]$,
allowing us more flexibility in evaluating the validity of a predictive set,
and we choose the loss itself to depend on user's queries, not only on their
label configuration $Y \in \mc{Y}$.  In many instances, users only care (and
therefore share information) about a fraction of their entire label
configuration: even if they have a full internal ranking of different bicycle
brands, they only reveal that they prefer (say) Specialized to
Cannondale when buying a new one;, in a segmentation
task, even if one can perfectly classify every pixel in an MRI scan between
healthy tissues and tumors, one probably cares more about a specific area
around a tumor.  The set of queries a user answers thus provides data about
their preferences, which the loss function should then reflect.  In
particular, one should not penalize a predictive set that contains
configurations matching all the probe functions that the user actually cares
about, even if it does not contain the correct full configuration itself.
By adapting our choice of loss to the probe functions $\{ \probe\}_{i \in
  \mc{I}_\mc{Y}}$ and hence to the specific partial supervision
task, we can efficiently construct predictive sets controlling the loss with
high probability~\eqref{eqn:partial-fpp-loss-quantile-coverage}.  We thus
consider user-specific losses $\{ \ell_{j}\}_{j=1}^{n+1}$ such that $\ell_j(C) =
0$ for all $C \subset \mc{Y}$ containing at least one element $y \in C$ such
that that $\probe(y) = \probe(Y_{j})$ for all $i \in I_{j}$, and construct a
predictive set mapping~$\what{C}_n$ such that
\begin{align}
\label{eqn:marginal-conformal-coverage-loss-weak-supervision}
\P\left( \ell_{n+1} (\what{C}_n(X_{n+1})) \le \delta \right) \ge 1-\alpha.
\end{align}

\newcommand{\nmpcerr}{\textup{Err}_\delta}

\section{Query-Adaptive Predictive Inference}
\label{sec:adap-conform-pred-query}
In this section, we define a false-discovery-type loss that leverages the collection of probe functions $\{ \probe \}_{i \in \mc{I}_\mc{Y}}$; we wish to produce interpretable predictive sets in terms of these queries.

\subsection{Probe-adapted predictive sets and loss}
\label{subsec:probe-adapted-pred-sets}
We first introduce definitions, focusing on structured label spaces
$\mc{Y}$ for which there exists a (fixed) collection of probe functions $\{
\probe \}_{i \in \mc{I}_\mc{Y}}$ with the identification
property~\eqref{eqn:probe-fcn-identifiable}.
\begin{definition}
\label{def:probe-function-set-definition}
For any $i \in \mc{I}_\mc{Y}$ and predictive set $C \subset \mc{Y}$,  define
\begin{align}
\label{eqn:probe-function-set-definition}
\probe(C) \defeq \begin{cases}
		 \varepsilon_i \in \{-1,1\} & \text{if} ~ \probe(y)=\varepsilon_i ~ \text{for all} ~ y \in C\\
           0 & \text{otherwise,}
		 \end{cases}
\end{align}
Additionally,  define $I(C) \defeq \{ i \in \mc{I}_\mc{Y} \mid \probe(C)\neq 0  \}$ to be the query indices for which $C$ makes a decision.
\end{definition}
Definition~\ref{def:probe-function-set-definition} extends the probe $\probe$ from $\mc{Y}$ to $2^{\mc{Y}}$ while ensuring that $\probe(\{y \}) = \probe(y)$ for all $y \in \mc{Y}$.
This extension to all subsets of $\mc{Y}$
allows us to characterize probe-adapted predictive sets
containing only configurations compatible with a subset of probe
functions.
\begin{definition}
A predictive set $C \subset \mc{Y}$ is \textup{probe-adapted} if
\begin{align}
\label{eqn:predictive-set-probe-adapted}
C = \left\{ y \in \mc{Y} \mid
~ \mbox{for~all~} i \in I(C), \, \probe(y) = \probe(C) \right\}.
\end{align}
\end{definition}
The index set $I(C)$ and probes $\{ \probe(C) \}_{i \in I(C)}$ of a probe-adapted predictive set $C$ entirely characterize its set of configurations, which is why we only use them to describe our predictive sets in the following.
A predictive set is not probe-adapted if it contains contradicting
configurations, e.g., in a ranking problem with binary comparisons, the
predictive set $C \defeq \{(1, 2, \dots, K), (K, K-1, \dots, 1) \}$ must have
$I(C) = \emptyset$, so that the set on the right in
equation~\eqref{eqn:predictive-set-probe-adapted} contains all
configurations.  $I(C)$ and $( \probe(C))_{i \in I(C)}$ fully
characterize a probe-adapted predictive set $C$, which presents
computational benefits: the space of configurations $\mc{Y}$ is
prohibitively large, but the number of probe functions itself is more
amenable.  For instance, in the ranking case, suppose that $C(x)$ is the set
of all configurations $y$ such that $y(1)=1$: listing all of them (there are
$(K-1)!$) is not a viable option, but 
it only takes the probe function $\varphi(y) = 2 \cdot \indic{y(1)=1} -1$ to describe the set.

We measure the informativeness of a predictive set $C$ through the size of the probe subset $I(C)$: the larger it is,  the more constrained any configuration in $C$ has to be. 
In contrast with previous work in conformal inference, we do not focus on the size $|C|$ of the predictive set,  as its connection with informativeness appears blurry. 
Indeed,  our ranking example with $y(1)=1$ shows a case where a predictive set containing $(K-1)!$ configurations provides the desired information
Our probe-adapt predictive sets should then control the following user dependent loss.
\begin{definition}
Given an instance $(Y,I) \in \mc{Y} \times 2^{\mc{I}_\mc{Y}}$ and a predictive set $C$,
the \emph{False Probe Proportion} (FPP) loss is
\begin{align}
\label{eqn:user-probe-loss}
\ell^p_{Y, I}(C) \defeq \dfrac{|\{i \in I \cap I(C) \mid \probe(Y) \neq \probe(C) \}|}{\left|I \cap I(C)\right| \vee 1}.
\end{align}
\end{definition}
The purpose of the FPP loss is straightforward: we get feedback on probe functions in $I$, while the predictive set $C$ abstains
on all but a subset $I(C)$ of the probes, so we compare the queries
in the intersection $I \cap I(C)$.  The overall loss simply is the
proportion of all probe functions $\{ \probe \}_{i \in I \cap I(C)}$ that do
not agree between the weak label $\weakset = \{ y \in \mc{Y} \mid \probe(y)
= \probe(Y) \text{ for all } i \in I\}$ and the predictive set $C$.  When the
intersection is empty, there is no possible discrimination, so the loss
is $0$.  Similarly to the initial coverage
guarantee~\eqref{eqn:marginal-conformal-coverage-full}, controlling the FPP
loss only provides a type 1 guarantee: it ensures that
the predictive set $C$ does not make too many false query decisions, but
does not prevent it from getting very large: in general, we expect a
trade-off between the size of $I(C)$ and the value of the probe loss.

\subsection{Conformalized query prediction}
\label{subsec:conformal-query-pred}

Our goal is to construct an informative probe-adapted confidence set mapping
$\what{C}_n : \mc{X} \toto \mc{Y}$ controlling the $1-\alpha$-value at risk
(VaR) of the loss at level $\delta \ge 0$, as in
Eq.~\eqref{eqn:marginal-conformal-coverage-loss-weak-supervision},
\begin{align*}
\P\left( \ell^p_{Y_{n+1}, I_{n+1}} (\what{C}_n(X_{n+1})) \le \delta \right) \ge 1-\alpha.
\end{align*}
We assume access to pre-trained individual query predictions, specifically, that there exists a collection of score functions 
\begin{align*}
\{ \score_i : \mc{X} \to \R \}_{i \in \mc{I}_\mc{Y}}.
\end{align*}
We call this set probe-based non-conformity scores; typically, we expect $\score_i(x)$ to be small for instances $x \in \mc{X}$ such that $P\left( \probe(Y) = 1 \mid X=x \right)$ is large, and vice-versa. 
An ideal score would be $\score_i^\text{oracle}(x) \defeq - \log P\left( \probe(Y) = 1 \mid X=x \right)$. 
While usually unattainable,  it will not impact the validity of the loss guarantee~\eqref{eqn:marginal-conformal-coverage-loss-weak-supervision}: as with fully supervised predictive inference methods,  we simply expect more accurate scores to produce more informative predictive sets.  
When the label space is already of the form $\{-1,1\}^K$, our query predictions are score functions $\{ s_k : \mc{X} \to \R \}_{k \in [K]}$ for each individual task $k \in [K]$,  but this assumption also applies to other structured prediction settings, e.g.\ rankings:

\begin{example}[Probe-based scores for ranking]
Suppose that $\mc{Y} = \biject_K$ and we have pairwise comparison probes $\varphi_{ij}(y) \defeq \sign( y^{-1}(j) - y^{-1}(i))$.
If we follow standard practice~\cite{CaoQiLiTsLi07, FreundIyScSi03} and train individual relevance functions~ $\{ r_k : \mc{X} \to \R \}_{k \in [K]}$, then a potential non-conformity score for $\varphi_{ij}$ is $\score_{ij}(x) \defeq r_j(x) - r_i(x)$. 
If $r_i(x)$ is larger than $r_j(x)$, we expect item $i$ to be more relevant than item $j$ and to have a higher rank in $x$'s preferences, i.e.\ $y^{-1}(i) < y^{-1}(j)$.
\end{example}

Leveraging nested conformal prediction~\cite{GuptaKuRa22, BatesAnLeMaJo21}, for each potential feature vector $x \in \mc{X}$, we consider probe-adapted predictive sets $\{ C^\score_\lambda(x) \}_{\lambda \ge 0}$ implicitly defined
by the constraints
\begin{align}
\label{eqn:threshold-pred-set}
I(C^\score_\lambda(x)) \defeq \{ i \in \mc{I}_\mc{Y} \mid \left|\score_i(x)\right| > \lambda \},  ~ \text{and} ~  \varphi_i(C^\score_\lambda(x)) = \sign(s_i(x)) ~ \text{for all} ~ i \in I(C^\score_\lambda(x)),
\end{align}
or more explicitly,
\begin{equation*}
  C^\score_\lambda(x) \defeq \left\{y \in \mc{Y} \mid \varphi_i(y) = \sign(s_i(x))
  ~ \mbox{for~all~} i ~ \mbox{s.t.}~ |s_i(x)| > \lambda
  \right\},
\end{equation*}
so that $C^\score_\lambda$ answers all probes $\probe$ for which it has
sufficient confidence in the output.  It is immediate that for all
$C_{\lambda_1}(x) \subset
C_{\lambda_2}(x)$ whenever $\lambda_1 \le \lambda_2$.
Additionally, if $s_i(x) \neq 0$ for all $i \in \mc{I}_\mc{\mc{Y}}$, then
$C_0(x)$ contains a single configuration by the identifiability
requirement~\eqref{eqn:probe-fcn-identifiable}, while $C_\lambda(x) =
\mc{Y}$ for any $\lambda > \max_{i \in \mc{I}_\mc{Y}} |s_i(x)|$.

\begin{remark}
We implicitly assume that the score functions are centered in the sense that $\score_i(x) = 0$ corresponds to the largest possible indecision. 
While this assumption simplifies our exposition and is reasonable for the models we study, we can also have two different thresholds $\lambda_{-} < \lambda_{+}$ such that
$
I(C_{\lambda_\pm}(x)) \defeq \{ i \in \mc{I}_\mc{Y} \mid \score_i(x) \le \lambda_- ~ \text{or} ~ \score_i(x) \ge \lambda_+ \},
$
which then requires strategies akin to~\citeauthor{AngelopoulosBaCaJoLe21}'s work~\citep{AngelopoulosBaCaJoLe21} to handle multiple parameters.
Algorithm~\ref{alg:fixed-step-sequence-partial-loss} to come allows such distinction.
\end{remark}

\subsubsection{Step-down weak conformal inference}
\begin{algorithm}
  \caption{Step-down probe-based conformalization}
  \label{alg:nested-uniform-probe-based-conformalization}
  \begin{algorithmic}
    \STATE {\bf Input:} sample $\{X_j,(\probe(Y))_{i \in I_j}
    \}_{j=1}^{n}$; probe-based score functions $\score_i: \mc{X} \to \R$ for $i \in \mc{I}_\mc{Y}$
    independent of the sample; desired quantile $1-\alpha \in (0,1)$, desired threshold $\delta > 0$; 

    \STATE For each $i \in [n]$, compute
    \begin{align}
      \label{eqn:uniform-non-conformity-score}
      \scorerv_j \defeq \inf \{ \lambda > 0 \mid \forall \lambda' \ge \lambda, \ell^p_{Y_j, I_j}(C^\score_{\lambda'}(X_j)) \le \delta \}.
    \end{align}

    \STATE Set 
    $\what{\lambda}_n^\text{down} \defeq (1+n^{-1}) (1-\alpha) \text{-quantile of} ~ \{ \scorerv_j  \}_{j=1}^n$.
    \STATE \textbf{Return:} probe-adapted predictive set mapping
    $\what{C}_n(x) \defeq C^\score_{\what{\lambda}_n^\text{down}}(x)$, satisfying
    \begin{align*}
    I(\what{C}_n(x)) =  \{ i \in \mc{I}_\mc{Y} \mid \left|\score_i(x)\right| > \what{\lambda}_n^\text{down} \},
    \end{align*}
    and 
    $
\varphi_i(\what{C}_n(x)) = \sign(s_i(x))
$ 
for all $i \in \mc{I}_\mc{Y}$.
  \end{algorithmic}
\end{algorithm}

Let us now relate the choice of threshold $\lambda \ge 0$ to the probe loss~\eqref{eqn:user-probe-loss}.
For an instance $(x,y,I) \in \mc{X} \times \mc{Y} \times 2^{\mc{I}_\mc{Y}}$, the predictive set $C_\lambda^\score(x)$ suffers a loss equal to
\begin{align*}
\dfrac{\sum_{i \in I} \indic{ -\probe(y)\cdot s_i(x) > \lambda}}{1 \vee \left(\sum_{i \in I} \indic{|s_i(x)| > \lambda}\right)} \in [0,1]
\end{align*}
An ideal threshold choice would be
\begin{align}
\label{eqn:lambda-up}
\lambda^\text{up} \defeq \min \{  \lambda \ge 0 \mid
 \P( \ell^p_{Y,I}(C_\lambda^\score(X)) \le \delta) \ge 1-\alpha \},
\end{align}
as it would maximize the informativeness of the predictive set while controlling the loss adequately.
However, unlike in standard conformal inference, the loss is not necessarily a decreasing function of $\lambda$.
One way to avoid the lack of monotonicity is to define the non-conformity scores as in equation~\eqref{eqn:uniform-non-conformity-score},  choosing $\lambda \ge 0$ with a step-down procedure,  ensuring that any $\lambda' \ge \lambda$ satisfies the same constraint.
The next theorem, whose proof we provide in Appendix~\ref{subsec:proof-thm-step-down-conform-validity} proves the validity of this first approach, which we summarize in Algorithm~\ref{alg:nested-uniform-probe-based-conformalization}.
\begin{theorem}
\label{thm:step-down-conform-validity}
Suppose that $(X_i, Y_i, I_i)_{i=1}^{n+1} \simiid P$. Then Algorithm~\ref{alg:nested-uniform-probe-based-conformalization} returns a probe-adapted predictive set mapping $\what{C}_n$ that controls the $1-\alpha$-VaR of the loss at level $\delta$:
\begin{align*}
\P\left( \ell^p_{Y_{n+1}, I_{n+1}} (\what{C}_n(X_{n+1})) \le \delta \right) \ge 1-\alpha.
\end{align*}
\end{theorem}
\begin{remark}
The computation of each score $\scorerv_j$ in equation~\eqref{eqn:uniform-non-conformity-score} requires $\mc{O}(|\mc{I}_\mc{Y}| \log |\mc{I}_\mc{Y}|)$ operations, as it amounts to sorting the vector $\left( \score_i(X_j)\right)_{i \in \mc{I}_\mc{Y}}$.
\end{remark}

\subsubsection{A less conservative step-up approach}
While the step-down procedure from Alg.~\ref{alg:nested-uniform-probe-based-conformalization} provably controls the $1-\alpha$-quantile of the loss,  it can (and,  in practice, often will) yield conservative predictive sets, in the sense that $\P\big( \ell^p_{Y_{n+1}, I_{n+1}} (\what{C}_n(X_{n+1})) \le \delta \big) \ll 1-\alpha$. 
Indeed, while we ideally wish to choose the threshold $\lambda^\text{up}$~\eqref{eqn:lambda-up}, 
in reality Alg.~\ref{alg:nested-uniform-probe-based-conformalization} targets the threshold
\begin{align}
\label{eqn:lambda-down}
\lambda^\text{down} \defeq \min \left\{ \lambda \ge 0: \P\left(\forall \lambda' \ge \lambda, ~ \ell^p_{Y_{n+1}, I_{n+1}} (C_{\lambda'}^\score(X_{n+1})) \le \delta \right) \ge 1-\alpha \right\},
\end{align}
which can be much larger when the function $\lambda \mapsto \ell^p_{Y_{n+1}, I_{n+1}} (C_{\lambda}^\score(X_{n+1}))$ is not decreasing.

\begin{algorithm}
  \caption{Step-up probe-based conformalization}
  \label{alg:step-up-probe-based-conformalization}
  \begin{algorithmic}
    \STATE {\bf Input:} sample $\{X_j,(\probe(Y))_{i \in I_j}
    \}_{j=1}^{n}$; probe-based score functions $\score_i: \mc{X} \to \R$ for $i \in \mc{I}_\mc{Y}$
    independent of the sample; desired quantile  $1-\alpha \in (0,1)$, desired threshold $\delta > 0$, tolerance $\varepsilon > 0$.

    \STATE For each $j \in [n]$, compute
    \begin{align}
      \label{eqn:uniform-non-conformity-score}
      \tilde \scorerv_j \defeq \inf \{ \lambda > 0 \mid \ell^p_{Y_j, I_j}(C^\score_{\lambda}(X_j)) \le \delta \}.
    \end{align}

    \STATE Set 
    $\what{\lambda}^\text{up}_n \defeq (1+n^{-1}) (1-\alpha) \text{-quantile of} ~ \{\tilde \scorerv_j  \}_{j=1}^n$.
    \STATE \textbf{Return:} probe-adapted predictive set mapping
    $\what{C}_n(x) \defeq C^\score_{\what{\lambda}^\text{up}_n + \varepsilon}(x)$ such that 
    \begin{align*}
    I(\what{C}_n(x)) =  \{ i \in \mc{I}_\mc{Y} \mid \left|\score_i(x)\right| > \what{\lambda}^\text{up}_n +\varepsilon  \},
    \end{align*}
    and 
    $
\varphi_i(\what{C}_n(x)) = \sign(s_i(x))
$ 
for all $i \in \mc{I}_\mc{Y}$.
  \end{algorithmic}
\end{algorithm}

Alg.~\ref{alg:step-up-probe-based-conformalization}, which is the step-up counterpart of Alg.~\ref{alg:nested-uniform-probe-based-conformalization}, provides the intuitive relaxation, but loses the type I error guarantee from Theorem~\ref{thm:step-down-conform-validity}.
A first alternative is to estimate the loss in coverage on some (potentially left out) fresh data, but Theorem~\ref{thm:step-up-conform-validity} below, which we prove in Appendix~\ref{subsec:proof-thm-step-up-conform-validity}, also relates the loss in coverage with how close to monotone the function $\lambda \mapsto \ell^p_{Y_{n+1}, I_{n+1}} (C_{\lambda}^\score(X_{n+1}))$ is.
Namely, for any threshold $\lambda \ge 0$ and tolerance level $\varepsilon >0$, let 
\begin{align*}
\nmpcerr(\lambda, \varepsilon) \defeq \P\left[ \text{There exists}~ \lambda' \in [0,\lambda] ~ \text{s.t.}~
\ell^p_{Y, I}(C^\score_{\lambda'}(X)) \le \delta < \ell^p_{Y, I}(C^\score_{\lambda+\varepsilon}(X))
\right],
\end{align*}
which measures the likelihood of simultaneously having a non $\delta$-valid predictive set $C^\score_{\lambda}(X)$ while there exists a  larger $\delta$-valid predictive set $C^\score_{\lambda'}(X)$. 
This probability is evidently $0$ when the function $\lambda \mapsto \ell^p_{Y, I}(C^\score_{\lambda}(X))$ is almost surely non-increasing; if it is not, then as long as the score function consistently ranks queries by increasing difficulty,  we expect $\nmpcerr(\lambda, \varepsilon)$ to remain small.
Our subsequent experiments suggest that typical score functions are consistent with this hypothesis.
The tolerance level $\varepsilon >0$ accounts for potential local perturbations of the random function $\lambda \mapsto \ell^p_{Y, I}(C^\score_{\lambda}(X))$: it is possible to have $\ell^p_{Y, I}(C^\score_{\lambda}(X))> \delta$ when $\ell^p_{Y, I}(C^\score_{\lambda'}(X)) \simeq \delta$ for some $\lambda' < \lambda$, but much less so that $\ell^p_{Y, I}(C^\score_{\lambda + \varepsilon}(X)) > \delta$.  

\begin{theorem}
\label{thm:step-up-conform-validity}
Suppose that $(X_i, Y_i, I_i)_{i=1}^{n+1} \simiid P$. Then Algorithm~\ref{alg:step-up-probe-based-conformalization} returns a probe-adapted predictive set mapping $\what{C}_n$ such that
\begin{align*}
\P\left( \ell^p_{Y_{n+1}, I_{n+1}} (\what{C}_n(X_{n+1})) \le \delta \right) \ge 1-\alpha - \E\left[ \nmpcerr(\what \lambda_n^\text{up}, \varepsilon) \right] .
\end{align*}
\end{theorem}
Theorem~\ref{thm:step-up-conform-validity} hints at two different ideas: on the one hand,  we expect most predictive models to yield a $1-\alpha$-VaR reasonably close to $\delta$, by controlling instead the $1-\alpha'$-VaR with $\alpha' \gtrsim \alpha$. 
On the other hand, $ \nmpcerr(\what \lambda_n^\text{up}, \varepsilon)$ is a quantity that depends on the algorithm, hence in practice we typically require additional data splitting to form an unbiased estimate.

\subsection{Multiple hypothesis testing query prediction}
In our setting, conformal inference based methods have two limitations. 
First, when dealing with non monotone losses,  they can either prove too conservative (Alg~\ref{alg:nested-uniform-probe-based-conformalization}) or require additional correction (Alg~\ref{alg:step-up-probe-based-conformalization}).
Additionally,  controlling the $1-\alpha$ VaR of FPP loss~\eqref{eqn:marginal-conformal-coverage-loss-weak-supervision} may be restrictive when the set $I$ of probe functions is small: the probe loss can only take few different values,
so being smaller than $\delta$ enforces it is $0$.
An expectation bound instead requires that, for some instance-dependent loss $\ell_{n+1}$,
\begin{align}
\label{eqn:marginal-conformal-coverage-loss-weak-supervision-expectation}
\E \left[ \ell_{n+1}\left( \what{C}_n(X_{n+1}) \right) \right] \le \delta,
\end{align}
but is not directly amenable to conformal inference methods.  Indeed, unlike
the variable $1\{\ell^p_{Y,I}(C^\score_\lambda(X)) \le \delta\} \in
\{0,1\}$, the random variable $\ell^p_{Y,I}(C^\score_\lambda(X))$ can take
many different values as a function of $\lambda \ge 0$, preventing
defining non-conformity scores as in
equation~\eqref{alg:nested-uniform-probe-based-conformalization}.
\begin{algorithm}
  \caption{Fixed Sequence Testing~\cite{AngelopoulosBaCaJoLe21} for weakly supervised predictive inference}
  \label{alg:fixed-step-sequence-partial-loss}
  \begin{algorithmic}
    \STATE {\bf Input:} sample $\{X_j,(\probe(Y))_{i \in I_j}
    \}_{j=1}^{n}$; probe-based score functions $\score_i: \mc{X} \to \R$ for $i \in \mc{I}_\mc{Y}$
    independent of the sample; desired expectation  $\delta \in (0,1)$,  probability of error $\alpha_\text{FST} \in (0,1)$,  grid space $\{0< \lambda_1 < \dots < \lambda_N \}$.

    \STATE For each $k\in [N]$, compute a $p$-value $p_k$ to test the hypothesis $H_k$~\eqref{eqn:fixed-seq-hypothesis}.

     \STATE Set 
    $\what{k}^\text{FST}_n \defeq \min\{k \mid \text{For all}~ k' \ge k, p_{k'} \le \alpha_\text{FST} \}$ and $\what{\lambda}^\text{FST}_n \defeq \lambda_{\what{k}^\text{FST}_n}$.
    
    \STATE \textbf{Return:} the probe-adapted predictive set mapping
    $\what{C}_n(x) \defeq C^\score_{\what{\lambda}^\text{FST}_n} (x)$ such that 
    \begin{align*}
    I(\what{C}_n(x)) =  \{ i \in \mc{I}_\mc{Y} \mid \left|\score_i(x)\right| > \what{\lambda}^\text{FST}_n  \},
    \end{align*}
    and 
    $
\varphi_i(\what{C}_n(x)) = \sign(s_i(x))
$ 
for all $i \in I(\what{C}_n(x))$.
  \end{algorithmic}
\end{algorithm}

An alternative is to make our problem fit~\citeauthor{AngelopoulosBaCaJoLe21}'s Fixed Sequence Testing~\cite{AngelopoulosBaCaJoLe21} approach, and instead discretize the parameter space $\R_+$ into an evenly spaced grid $(\lambda_1 , \dots  \lambda_N)$, with associated individual $p$-values $\{ p_k \}_{k=1}^N$, where each of them tests the hypothesis
\begin{align}
\label{eqn:fixed-seq-hypothesis}
H_k: \E \left[ \ell_{n+1}\left( C_{\lambda_k}(X_{n+1}) \right) \right] > \delta,
\end{align}
using the partially supervised calibration set $\{X_j,  (\probe(Y_j))_{i \in I_j} \}_{j=1}^n$.
Proposition~\ref{prop:hoe-bent-ineq} below provides an example of valid
$p$-values for this hypothesis, using the Hoeffding-Bentkus inequality~\cite{Bentkus04, Hoeffding63}. 
The key here is that, while the FPP loss is not necessarily instance-wise monotone, our potential predictive sets still have a built-in order: they are nested, and we expect to reject the hypothesis $H_k$~\eqref{eqn:fixed-seq-hypothesis} more easily for higher values of $\lambda_k$.

\begin{proposition}[Hoeffding-Bentkus inequality~\cite{BatesAnLeMaJo21}]
\label{prop:hoe-bent-ineq}
Let $h(a,b) \defeq a \log\frac{a}{b} + (1-a) \log \frac{1-a}{1-b}$, and $\bar L_k \defeq \frac{1}{n}\sum_{j=1}^n \ell_j(C_{\lambda_k}(X_j))$, and assume that $\{ (X_j, Y_j, I_j) \}_{j=1}^{n+1} \simiid P$. Then, 
\begin{align*}
p_k^\text{HB} \defeq \min\left\{ \exp\left( -n h(\bar L_k \wedge \delta, \delta) \right), e \P \left( \mathrm{Bin}(n,\delta) \le \left\lceil n\bar L_k \right\rceil \right) \right\}
\end{align*}
is a valid p-value for hypothesis $H_k$~\eqref{eqn:fixed-seq-hypothesis}.
\end{proposition}

Algorithm~\ref{alg:fixed-step-sequence-partial-loss} summarizes this fixed sequence procedure,  and basically continues exploring the grid down as long as we reject each hypothesis. 
When the grid is coarse enough, we expect this procedure to return $\what{\lambda}^\text{FST}_n$ close to
\begin{align}
\label{eqn:lambda-fst}
\lambda^\text{FST} \defeq \inf\{ \lambda > 0 \mid \text{For all}~ \lambda' \ge \lambda, ~ \E \left[ \ell_{n+1}\left( C_{\lambda'}(X_{n+1}) \right) \right] \le \delta \}.
\end{align}
which, similarly to Alg.~\ref{alg:nested-uniform-probe-based-conformalization},  can be conservative if the expectation is not a decreasing function of $\lambda$. 
However, this similarity hides a key difference: while the (random) loss function itself may not be monotone,  in many applications, we expect its expectation of the loss to be, hence we expect $\lambda^\text{FST}$~\eqref{eqn:lambda-fst} to be smaller than $\lambda^\text{down}$~\eqref{eqn:lambda-down}.
Notably, Alg.~\ref{alg:fixed-step-sequence-partial-loss} controls the expectation of future losses with high probability over the calibration set. 

\begin{theorem}
\label{thm:fst-validity}
Suppose that the $p$-values $\{ p_k \}_{k=1}^N$ in Alg.~\ref{alg:fixed-step-sequence-partial-loss} are valid, i.e.\ that under $H_k$, we have $\P(p_k \le u) \le u$ for all $u \in [0,1]$,  then Alg.~\ref{alg:fixed-step-sequence-partial-loss} returns a predictive set mapping $\what{C}_n$ that satisfies
\begin{align*}
\P\left( \E \left[ \ell_{n+1}\left( \what{C}_n(X_{n+1}) \right) \mid  \what{C}_n \right] \le \delta \right) \ge 1- \alpha_\text{FST}.
\end{align*}
\end{theorem}
\noindent As the proof is essentially identical to that of~\cite[Proposition
  4]{AngelopoulosBaCaJoLe21}, we defer it to
Appendix~\ref{subsec:proof-thm-fst-validity}.

If one desires a quantile-type guarantee on the
loss~\eqref{eqn:marginal-conformal-coverage-loss-weak-supervision}, one can
replace the hypothesis~\eqref{eqn:fixed-seq-hypothesis} by
\begin{align*}
H_{k, \alpha} ~ : ~ \P\left[ \ell_{n+1}\left( C_{\lambda_k}(X_{n+1}) \right) > \delta\right] > \alpha, 
\end{align*}
which replaces each loss
$\ell^p_{Y,I}(C^\score_\lambda(X))$ by $1\{\ell^p_{Y,I}(C^\score_\lambda(X)) > \delta\}$ and $\delta$ by $\alpha$ in Proposition~\ref{prop:hoe-bent-ineq}.
Theorem~\ref{thm:fst-validity} then guarantees that
\begin{align*}
\P\left( \P \left[ \ell_{n+1}\left( \what{C}_n(X_{n+1}) \right)  > \delta \mid  \what{C}_n \right] \le \alpha \right) \ge 1- \alpha_\text{FST},
\end{align*}
so with high probability,
at most an $\alpha$-fraction of test examples have loss above $\delta$.

\section{Probe conditional predictive sets}
A recurring question~\cite{Vovk12, RomanoPaCa19, BarberCaRaTi19a, RomanoSeCa20} in conformal inference is the feature-conditional validity of the predictive sets: guarantees we can provide on the loss conditional on $X_{n+1} = x$ (which, unfortunately, are limited~\cite{BarberCaRaTi19a}).
In this section, we ask a related but distinct question.
Here, for each $X$ there exists a set $I \subset \mc{I}_\mc{Y}$ of queries
for which we wish to provide valid coverage.
A natural goal is to target coverage conditional on
the index set: we seek
\begin{align}
  \label{eqn:ind-conditional-coverage}
  \E\left[
    \ell^p_{Y_{n+1}, I_{n+1}} (\what{C}_n(X_{n+1})) \mid I_{n+1} = I \right]
  \le \delta
\end{align}
for all $I \subset \mc{I}_\mc{Y}$.  As are likely easier to answer than
others---think, for example, the top item versus the thirty-fifth item in a
ranking task---we expect the distribution of the loss to vary over choices
$I$.  The coverage~\eqref{eqn:ind-conditional-coverage} is challenging to
achieve for the same reasons that conditional coverage
is~\cite[cf.][]{BarberCaRaTi19a}.
A naive work-around is to construct
predictive sets for each $I \subset \mc{I}_{\mc{Y}}$,
$\{\what{C}_{n, I}(X_{n+1})\}_{I \in
  \mc{I}_\mc{Y}}$, each valid  when
$I_{n+1} = I$, but to do so would require
multiple observations
$I_j = I$ for each $I \subset \mc{I}_\mc{Y}$ in the calibration set, which
is unrealistic.  Instead, we propose a heuristic that still preserves the
initial guarantees from our three conformalization methods, but
produces predictive sets with losses uniformly closer to
$\delta$ under appropriate data generating distributions.


Our idea consists in modifying the predictive set sequence $\{ C_\lambda^\score \}_{\lambda \ge 0}$~\eqref{eqn:threshold-pred-set} to allow instance dependent thresholds. 
For each potential query $i \in \mc{I}_\mc{Y}$, we assume access to two different objects, namely
(i) a predictor $\what{\probe} : \mc{X} \to \{-1,1\}$, and
(ii) an estimate $\what{\pi}_i : \mc{X} \to [0,1]$ of the query accuracy $\pi_i(x) \defeq \P( \probe(Y) = \what{\probe}(X) \mid X=x)$.
In our initial set-up with scoring functions $\{ \score_i \}_{i \in \mc{I}_\mc{Y}}$,  this would correspond to $\what{\probe}(x) \defeq \sign(\score_i(x))$ and $\what{\pi}_i(x) = \psi(|\score_i(x)|)$ for some increasing link function $\psi : \R_+ \to [0,1]$.
We  associate the pair $(\what{\varphi}, \what{\pi})$ to probe-adapted predictive set mappings $\{C_\eta \}_{\eta \in [0,1]}$ defined
implicitly (as in Eq.~\eqref{eqn:threshold-pred-set}) via
\begin{align}
\label{eqn:probe-adapted-eta-threshold}
I(C_\eta(x)) \defeq \{ i \in \mc{I}_\mc{Y} \mid \what{\pi}_i(x) > \eta \}, ~ \text{and} ~ \probe(C_\eta(x))= \what{\probe}(x).
\end{align}

We develop an adaptive threshold via the ansatz that conditional on $X = x$,
individual query mistakes $\{\probe(Y) \neq \what{\probe}(X)\}_{i \in I}$
are independent of $I \subset \mc{I}_\mc{Y}$ (i.e., given $x$, whether or
not we make a mistake for query $i$ is independent of the choice of relevant
queries).  In that case, the FPP loss of the predictive set $C_\eta(x)$ is
the mean of (possibly dependent) Bernoulli variables with
probabilities of success $\{1-\pi_i(x) \}_{i \in I, \what{\pi}_i(x) >
  \eta}$, so it is in theory straightforward to tune $\eta \in [0,1]$
by setting
\begin{align}
  \eta^\star(x, \delta) &\defeq \inf\left\{ \eta \in [0,1] \mid 
  \E[ \ell^p_{Y, I} \left( C_{\eta}(X)]
  \mid I, X = x \right) \le 1-\delta \right\} \notag \\
  \label{eqn:adaptive-threshold-choice}
  & \,=\inf\bigg\{ \eta \in [0,1] \mid
  \frac{1}{|\{ i \in I: \pi_i(x) > \eta \}|}
  \sum_{i \in I: \pi_i(x) > \eta} (1-\pi_i(x))  \le 1-\delta \bigg\}.
\end{align}
Alg.~\ref{alg:bernoulli-pred-set} materializes this idea
to produce a nested sequence $\{ C_{\eta^\star(\cdot,  \delta)}(\cdot) \}_{0\le \delta \le 1}$ of predictive sets.
While attractive, these predictive sets come with two caveats. 
First, they require the set of relevant queries $I_{n+1}$ of any new $X_{n+1} \in \mc{X}$ \emph{a priori} to compute the adaptive threshold. 
Second, real datasets are unlikely to satisfy the independence assumption: in practice, one would expect queries to exhibit some specific structure so that errors have some correlation.

However, Alg.~\ref{alg:bernoulli-pred-set} returns nested confidence sets $\{C_{\eta^\star(x, \delta)}(x)\}_{0 \le \delta \le 1}$, to which one can also apply
  Alg.~\ref{alg:fixed-step-sequence-partial-loss} (or another conformalization algorithm):
 one then obtains a $\what{\delta}_n$---where $\delta$ basically functions as $\lambda$ in the algorithms---, and one thus adaptively chooses a ``nominal'' $\delta$ in the construction~\eqref{eqn:adaptive-threshold-choice} of $\eta\opt$ to then provide coverage at the target level $\delta$.
  
\begin{algorithm}
  \caption{Bernoulli sequence predictive set}
  \label{alg:bernoulli-pred-set}
  \begin{algorithmic}
    \STATE {\bf Input:} sample $\{X, I \}$; predictor $\what{\probe} : \mc{X} \to \{-1,1\}$ and error estimate $\what{\pi} : \mc{X} \to [0,1]$
    independent of the sample; desired expected loss  $1-\delta \in (0,1)$.

    \STATE Sort  confidence probabilities $\what{\pi}_{i_1}(X) \ge \dots \ge \what{\pi}_{i_N}(X)$, with $I = \{ i_j \mid 1\le j \le N\} \subset \mc{I}_\mc{Y}$.

    \STATE Set
    \begin{align*}
    J(x, \delta) \defeq \max \left\{ 1\le J \le N \mid  \frac{1}{J}\sum_{j=1}^J \what{\pi}_{i_j}(x) \ge \delta \right\}, 
    \end{align*}
    with the convention $\max(\emptyset) = 0$, and 
	\begin{align*}
	\eta^\star(x, \delta) \defeq
	\begin{cases}
	\what{\pi}_{i_{J(x, \delta)+1}}(x) ~ &\text{if}  ~ J(x, \delta)<N \\
	0 ~&\text{if} ~ J(x, \delta) = N.
	\end{cases}
\end{align*}	    
    
    \STATE \textbf{Return:} the probe-adapted predictive set
    $C_{\eta^\star(x,\delta)}(x)$.
  \end{algorithmic}
\end{algorithm}
%

\section{Experiments}
\label{sec:experiments}
We present a comparison of our different weakly supervised
procedures for a ranking task and a hierarchical classification task.

\subsection{Ranking documents by query relevance}

In this experiment, we wish to rank a set of potential documents by order of
relevance to a user query.  A typical example is
a search engine: a user performs a search, and the task is to sort Web
pages that best answer that query among a set of
potential pages.  We thus experiment with
the Microsoft LETOR data set~\cite{QiniLi13}.  For each potential query/document
pair $(x,z)$, the dataset aggregates several quantities of interest to
determine whether $d$ is relevant to $x$ into a $d=46$-dimensional feature
vector $\phi(x, z) \in \R^d$.  We associate each query $x$ to a set of
potentially relevant documents $Z(x) \defeq \{ z_j \}_{j=1}^{|D(x)|}$, as
well as a full ranking $Y \in \biject_{|Z(x)|}$ that orders these documents
according to their relevance.  Our goal is to retrieve that ranking using
the feature vectors $ \{ \phi(x, z_j) \}_{j=1}^{Z(x)}$.

Following the same procedure as in~\cite{CauchoisGuAlDu22}, we train relevance scores $r(x,z)$ that measure the saliency of a document $z$ to query $x$.
We additionally explain in Appendix~\ref{subsec:appendix-letor} how we choose $\choose{Z(x)}{2}$ possible pairs of documents as user queries to introduce weak supervision.

\begin{figure}[ht]
 \centering
  \begin{overpic}[
  				scale=0.5]{%
     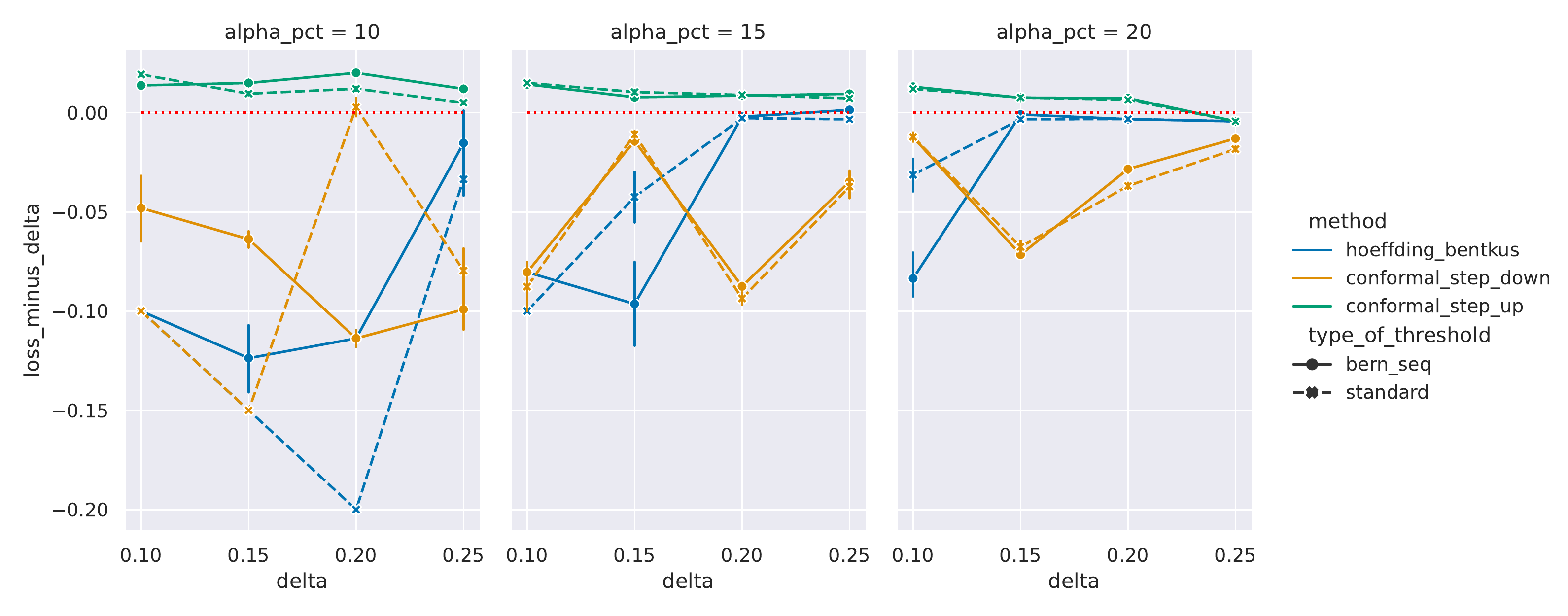}
     
    \put(15, 0){
      \tikz{\path[draw=white, fill=white] (0, 0) rectangle (4cm, .5cm)}
    }
    
        \put(40, 0){
      \tikz{\path[draw=white, fill=white] (0, 0) rectangle (4cm, .5cm)}
    }
    
        \put(62, 0){
      \tikz{\path[draw=white, fill=white] (0, 0) rectangle (4cm, .5cm)}
    }
    
    \put(17, 0){
       \small $\delta$
    }
        \put(42, 0){
       \small $\delta$
    }
        \put(67, 0){
       \small $\delta$
    }

        \put(10, 36.3){
      \tikz{\path[draw=white, fill=white] (0, 0) rectangle (4cm, .5cm)}
    }
    
        \put(32, 36.3){
      \tikz{\path[draw=white, fill=white] (0, 0) rectangle (4cm, .5cm)}
    }
    
        \put(60, 36.3){
      \tikz{\path[draw=white, fill=white] (0, 0) rectangle (4cm, .5cm)}
    }
    
    \put(15, 37){
       \small $\alpha = 10\%$
    }
        \put(40, 37){
       \small $\alpha = 15\%$
    }
        \put(65, 37){
       \small $\alpha = 20\%$
    }
    
    \put(70, 0){
      \tikz{\path[draw=white, fill=white] (0, 0) rectangle (4cm, .3cm)}
    }

    \put(0, 13){
      \tikz{\path[draw=white, fill=white] (0, 0) rectangle (.4cm, 4cm)}
    }
    \put(0, 10){\rotatebox{90}{
        \small $\quantile_{1-\alpha}\left( \ell^p_{Y,I}( \what{C}(X) ) \right) - \delta$}}
    
         \put(82, 24.5){
      \tikz{\path[draw=white, fill=white] (0, 0) rectangle (4cm, .5cm)}
    }
         \put(85, 23){
      \tikz{\path[draw=white, fill=white] (0, 0) rectangle (4cm, .4cm)}
    }
         \put(85, 21){
      \tikz{\path[draw=white, fill=white] (0, 0) rectangle (4cm, .4cm)}
    }
         \put(85, 19){
      \tikz{\path[draw=white, fill=white] (0, 0) rectangle (4cm, .4cm)}
    }
    
         \put(82, 24.5){
      \small \textbf{Method}
    }
         \put(85, 23){
      \scriptsize Alg.~\ref{alg:fixed-step-sequence-partial-loss}
    }
         \put(85, 21){
      \scriptsize Alg.~\ref{alg:nested-uniform-probe-based-conformalization}
    }
         \put(85, 19){
      \scriptsize Alg.~\ref{alg:step-up-probe-based-conformalization}
    }

         \put(82, 16.7){
      \tikz{\path[draw=white, fill=white] (0, 0) rectangle (4cm, .3cm)}
    }
         \put(85, 14.7){
      \tikz{\path[draw=white, fill=white] (0, 0) rectangle (4cm, .4cm)}
    }
         \put(85, 13.8){
      \tikz{\path[draw=white, fill=white] (0, 0) rectangle (4cm, .4cm)}
    }
    
             \put(82, 17.6){
      \small \textbf{Predictive set sequence}
    }
         \put(85, 15.8){
      \scriptsize $\{ C_{\eta^\star(X,\delta)}(X) \}_{0 \le \delta \le 1}$~\eqref{eqn:adaptive-threshold-choice}
    }
         \put(85, 13.6){
      \scriptsize $\{ C_\lambda^\score(X) \}_{\lambda \ge 0}$~\eqref{eqn:threshold-pred-set}
      }
      
  \end{overpic}
  \caption{Results for LETOR ranking dataset~\cite{QiniLi13}. 
 Difference between the $1-\alpha$-quantile on the test set of the loss $\ell^p_{Y,I}(C(X))$~\eqref{eqn:user-probe-loss} and $\delta$,
  for different, $\alpha$, $\delta$, and  methods.
  We average our results over $10$ runs, with error bars displaying the interquartile range.
  A predictive set mapping $C$ provides adequate control of the loss if that difference falls below the $y=0$ line, as we then have $\P(\ell^p_{Y,I}( C(X)) > \delta) \le \alpha$.
  }
  \label{fig:letor-quantile-loss}
\end{figure}

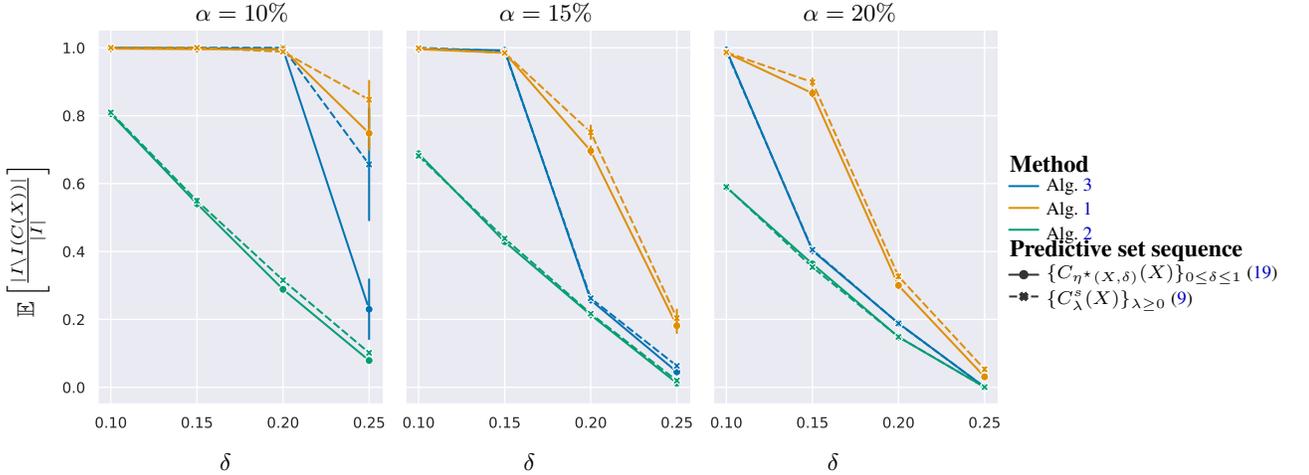
\begin{figure}[h!]
 \centering
  \begin{overpic}[
  				scale=0.50]{%
     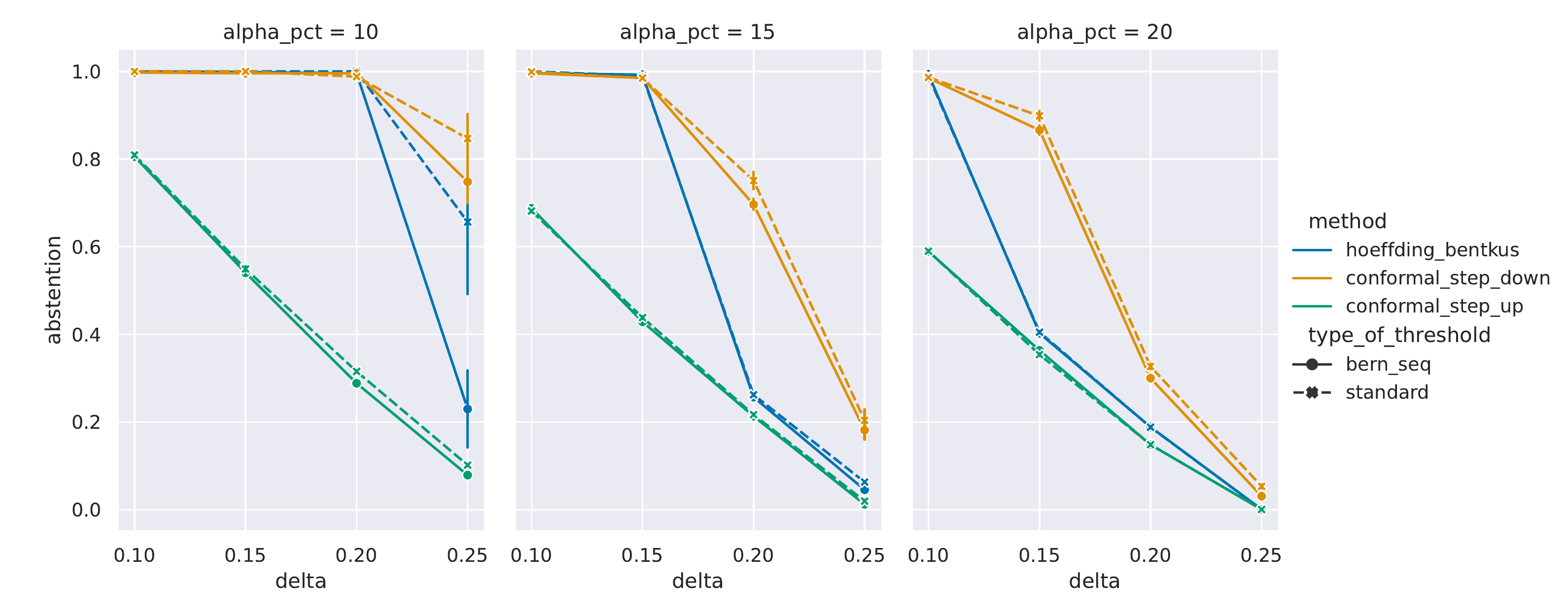}
     
    \put(15, 0){
      \tikz{\path[draw=white, fill=white] (0, 0) rectangle (4cm, .5cm)}
    }
    
        \put(40, 0){
      \tikz{\path[draw=white, fill=white] (0, 0) rectangle (4cm, .5cm)}
    }
    
        \put(62, 0){
      \tikz{\path[draw=white, fill=white] (0, 0) rectangle (4cm, .5cm)}
    }
    
    \put(17, 0){
       \small $\delta$
    }
        \put(42, 0){
       \small $\delta$
    }
        \put(67, 0){
       \small $\delta$
    }

        \put(10, 36.3){
      \tikz{\path[draw=white, fill=white] (0, 0) rectangle (4cm, .5cm)}
    }
    
        \put(32, 36.3){
      \tikz{\path[draw=white, fill=white] (0, 0) rectangle (4cm, .5cm)}
    }
    
        \put(60, 36.3){
      \tikz{\path[draw=white, fill=white] (0, 0) rectangle (4cm, .5cm)}
    }
    
    \put(15, 37){
       \small $\alpha = 10\%$
    }
        \put(40, 37){
       \small $\alpha = 15\%$
    }
        \put(65, 37){
       \small $\alpha = 20\%$
    }
    
    \put(70, 0){
      \tikz{\path[draw=white, fill=white] (0, 0) rectangle (4cm, .3cm)}
    }

    \put(0, 13){
      \tikz{\path[draw=white, fill=white] (0, 0) rectangle (.6cm, 5cm)}
    }
    \put(0, 12){\rotatebox{90}{
        \small $\E \left[ \frac{\left| I \setminus I(C(X)) \right|}{|I|} \right] $}}

         \put(82, 24.5){
      \tikz{\path[draw=white, fill=white] (0, 0) rectangle (4cm, .5cm)}
    }
         \put(85, 23){
      \tikz{\path[draw=white, fill=white] (0, 0) rectangle (4cm, .4cm)}
    }
         \put(85, 21){
      \tikz{\path[draw=white, fill=white] (0, 0) rectangle (4cm, .4cm)}
    }
         \put(85, 19){
      \tikz{\path[draw=white, fill=white] (0, 0) rectangle (4cm, .4cm)}
    }
    
         \put(82, 24.5){
      \small \textbf{Method}
    }
         \put(85, 23){
      \scriptsize Alg.~\ref{alg:fixed-step-sequence-partial-loss}
    }
         \put(85, 21){
      \scriptsize Alg.~\ref{alg:nested-uniform-probe-based-conformalization}
    }
         \put(85, 19){
      \scriptsize Alg.~\ref{alg:step-up-probe-based-conformalization}
    }

         \put(82, 16.7){
      \tikz{\path[draw=white, fill=white] (0, 0) rectangle (4cm, .3cm)}
    }
         \put(85, 14.7){
      \tikz{\path[draw=white, fill=white] (0, 0) rectangle (4cm, .4cm)}
    }
         \put(85, 13.8){
      \tikz{\path[draw=white, fill=white] (0, 0) rectangle (4cm, .4cm)}
    }
    
             \put(82, 17.6){
      \small \textbf{Predictive set sequence}
    }
         \put(85, 15.8){
      \scriptsize $\{ C_{\eta^\star(X,\delta)}(X) \}_{0 \le \delta \le 1}$~\eqref{eqn:adaptive-threshold-choice}
    }
         \put(85, 13.6){
      \scriptsize $\{ C_\lambda^\score(X) \}_{\lambda \ge 0}$~\eqref{eqn:threshold-pred-set}
      }
     
  \end{overpic}
  \caption{Results for LETOR ranking dataset~\cite{QiniLi13}. 
   Average abstention $1-\dfrac{|I \cap I(\what{C}_n(X)) | }{|I|}$ on the test set for different values of $\alpha$ and $\delta$ and respective methods.
  Lower abstention is better, since more informative.
  We average our results over $10$ runs, with error bars displaying the interquartile range.
  }
  \label{fig:letor-abstention-avg}
\end{figure}

\paragraph{Experimental results}
We run three conformalization procedures (Algs.~\ref{alg:nested-uniform-probe-based-conformalization}, ~\ref{alg:step-up-probe-based-conformalization}, and~\ref{alg:fixed-step-sequence-partial-loss}) with either predictive set sequence $\{ C^\score_\lambda \}_{\lambda \ge 0}$~\eqref{eqn:predictive-set-probe-adapted} and $\{ C_{\eta^\star(\cdot,  \delta)}(\cdot) \}_{0\le \delta \le 1}$~\eqref{eqn:adaptive-threshold-choice} as inputs, and present results for these six different procedures.
Our goal here is to construct a confidence set mapping $\what{C}_n$ satisfying equation~\eqref{eqn:marginal-conformal-coverage-loss-weak-supervision},controlling the $1-\alpha$ quantile of the loss at level $\delta$.
We vary $\delta$ and $\alpha$ across a range of parameters and present our results in Figures~\ref{fig:letor-quantile-loss} and~\ref{fig:letor-abstention-avg} .
Our results fit our initial expectations: on Figure~\ref{fig:letor-quantile-loss},  it is clear that Alg.~\ref{alg:nested-uniform-probe-based-conformalization} produces predictive sets that are in general too conservative, although not as much if we construct the initial nested sequence of sets with Alg.~\ref{alg:bernoulli-pred-set}.
On the other hand, Alg.~\ref{alg:step-up-probe-based-conformalization} is slightly too optimistic, and the $1-\alpha$ quantile of the loss is consistently a few percent above $\delta$, as Theorem~\ref{thm:step-up-conform-validity} predicts. 
That said, as we could expect, the amount by which it exceeds $\delta$ remains small, so much that our estimate remains within the margin of error.
We run Alg.~\ref{alg:fixed-step-sequence-partial-loss} with $\alpha_{\text{FST}} = 10\%$, which produces predictive sets with a fraction $1-\alpha$  of predictions consistently below the error threshold $\delta$, and often close to it, except for small values of $\alpha$ and $\delta$, corresponding to very stringent predictive sets; in this case, the initial scoring model is simply not good enough to produce predictions with the required confidence. 
We provide additional figures and details in Appendix~\ref{subsec:appendix-letor}.

\subsection{Tree structured prediction on ImageNet~\cite{DengDoSoLiLiFe09}}

We now present the results of an experiment dealing with tree-structured classification, specifically with the ImageNet benchmark image classification task. 
In such tasks, fine-grained labels (leaves of the tree) are harder to predict than more general classes (internal nodes of the tree): even if the predictive model aims to predict the leaf label, we can return a coarser label (i.e.\ a sub-tree) if the desired type I error is too small to only return a single label.
Originating with the WordNet hierachy, the $K=1000$ labels from the ImageNet data set form a hierarchical prediction problem, in which every inner node of the label tree represents a more general class of objects, e.g.\ physical entities, animals, mammals, dogs,etc... 
In this context,  weak supervision arises when a labeler provides an internal label (e.g.\ dog) rather than a leaf (e.g.\ German Sheperd), or a negative label (e.g.\ not a cat).
The label space is the set of leaves $\mc{Y}$ of a directed tree $\mc{T}=(\mc{V}, \mc{E})$, for which every node $v \in \mc{V}$ of the tree corresponds to the query
\begin{align*}
\varphi_v(y) = 2 \cdot \indic{v~\text{is an ancestor of}~y~\text{in}~\mc{T}}-1.
\end{align*}

\paragraph{Motivation}
\begin{figure}
 \centering
  \begin{overpic}[
  				scale=0.9]{%
     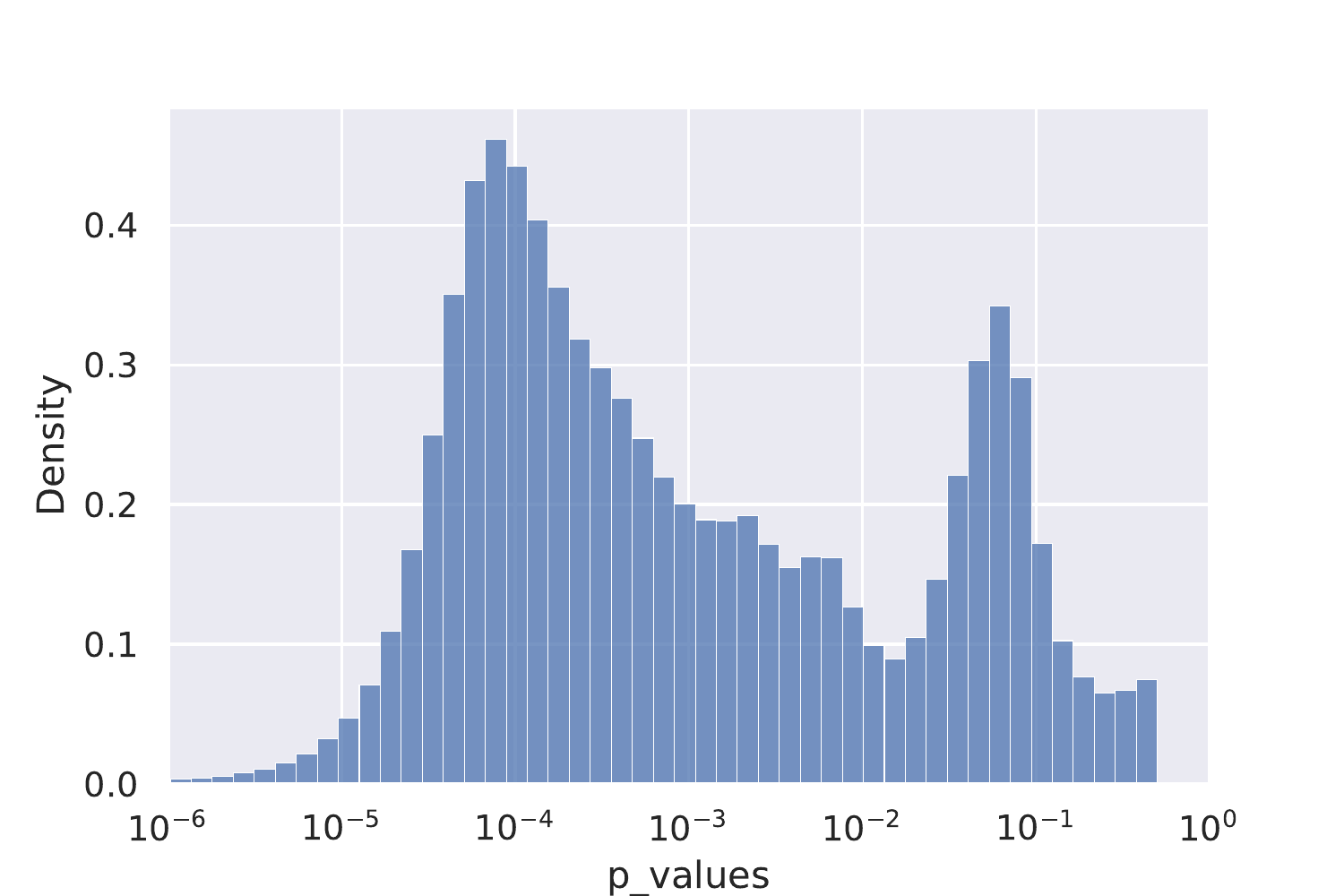}
             \put(40, -0.5){
      \tikz{\path[draw=white, fill=white] (0, 0) rectangle (5cm, .5cm)}
    }
    
    \put(46, 1){
        $\what{\pi}_i(x)$
    }
    
     \put(2, 20){
      \tikz{\path[draw=white, fill=white] (0, 0) rectangle (.5cm, 4cm)}
    }
    \put(3, 28){\rotatebox{90}{
         Density}}
  \end{overpic}
  \caption{Distribution of the error estimator $\what{\pi}_i(x)$ on queries from the ImageNet dataset.  
  }
  \label{fig:imagenet-pvalues}
\end{figure}
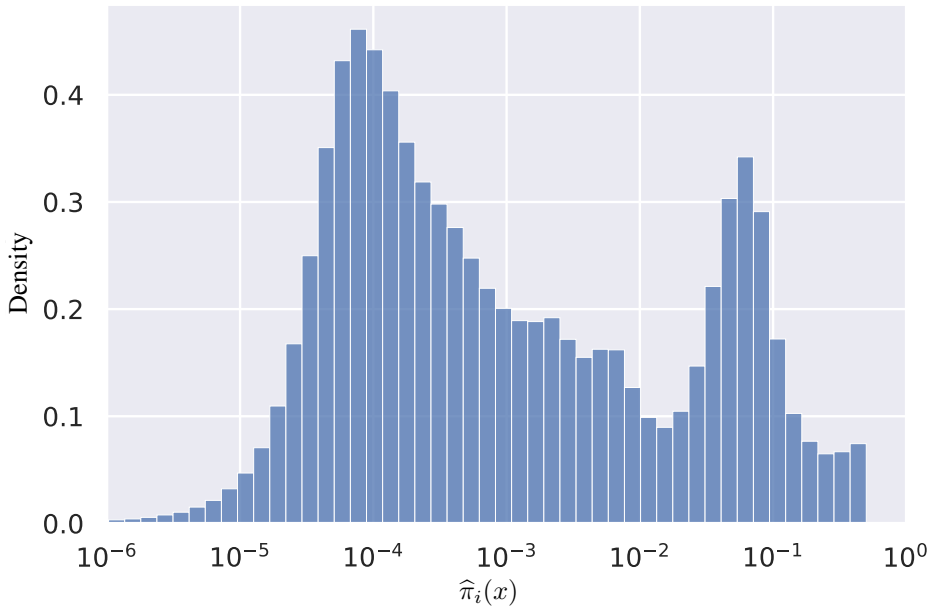

This experiment exemplifies prediction tasks where we expect heterogeneous confidences in our probes predictions. 
Indeed, we typically expect our predictive model to answer queries higher up in the tree---far from the label---with high accuracy, but to show more confusion as it gets closer to the label, $y$ e.g.\ between different breeds of dogs.
Figure~\ref{fig:imagenet-pvalues} suggests precisely this: while a majority of queries have error probability $<.001$ (high confidence), but a significant fraction has lower confidence (around $.1$), suggesting a difficult subset of queries.

While these error estimates are model-based, and in practice over-confident,  it also suggests that predictive sets may benefit from a data instance threshold as in equation~\eqref{eqn:adaptive-threshold-choice}: we expect instances with more queries on the left side of the distribution to control the loss even with lower threshold values, whereas we probably need to be more conservative for instances with uncertain queries.

We provide experimental details in Appendix~\ref{subsec:appendix-imagenet}, and in particular describe how we introduce weak supervision in the dataset by mostly sampling queries close to the true label in the tree.

\begin{figure}
 \centering
  \begin{overpic}[
  				scale=0.5]{%
     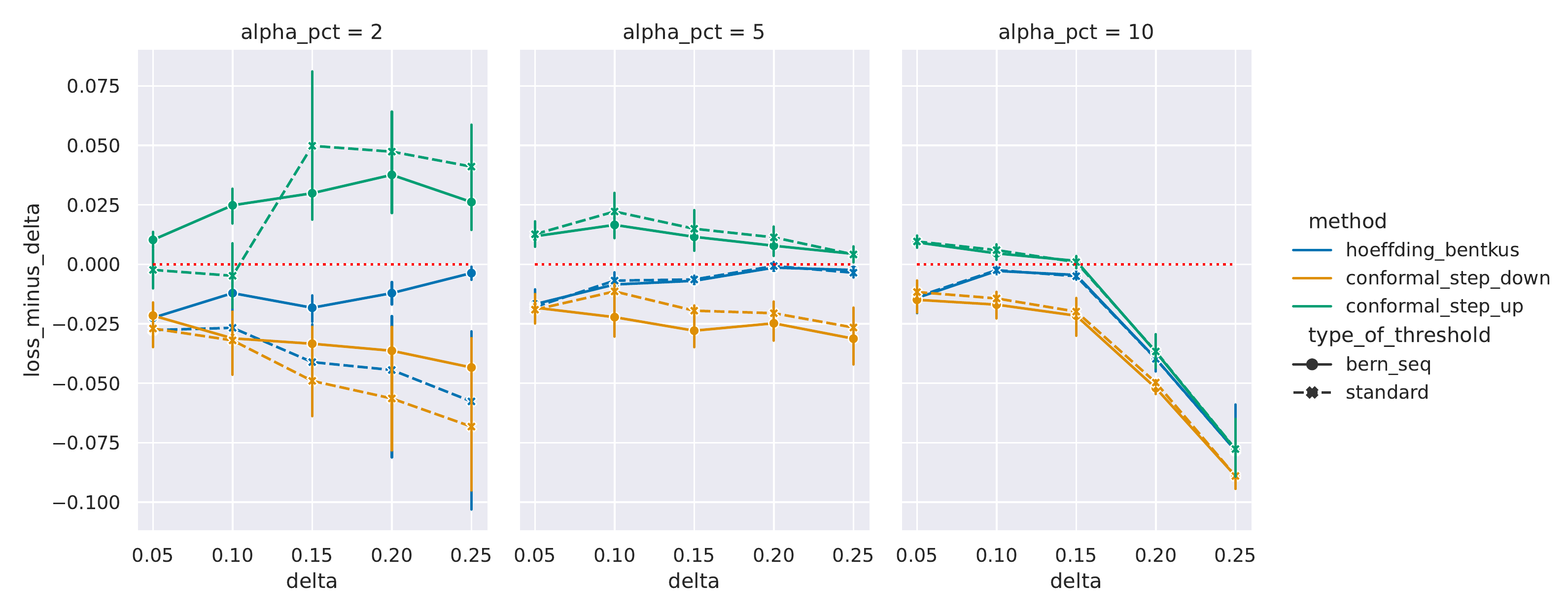}
        \put(15, 0){
      \tikz{\path[draw=white, fill=white] (0, 0) rectangle (4cm, .5cm)}
    }
    
        \put(40, 0){
      \tikz{\path[draw=white, fill=white] (0, 0) rectangle (4cm, .5cm)}
    }
    
        \put(62, 0){
      \tikz{\path[draw=white, fill=white] (0, 0) rectangle (4cm, .5cm)}
    }
    
    \put(17, 0){
       \small $\delta$
    }
        \put(42, 0){
       \small $\delta$
    }
        \put(67, 0){
       \small $\delta$
    }

        \put(10, 36.3){
      \tikz{\path[draw=white, fill=white] (0, 0) rectangle (4cm, .5cm)}
    }
    
        \put(32, 36.3){
      \tikz{\path[draw=white, fill=white] (0, 0) rectangle (4cm, .5cm)}
    }
    
        \put(60, 36.3){
      \tikz{\path[draw=white, fill=white] (0, 0) rectangle (4cm, .5cm)}
    }
    
    \put(15, 37){
       \small $\alpha = 2\%$
    }
        \put(40, 37){
       \small $\alpha = 5\%$
    }
        \put(65, 37){
       \small $\alpha = 10\%$
    }
    
    \put(70, 0){
      \tikz{\path[draw=white, fill=white] (0, 0) rectangle (4cm, .3cm)}
    }

    \put(0, 13){
      \tikz{\path[draw=white, fill=white] (0, 0) rectangle (.4cm, 5cm)}
    }
    \put(0, 10){\rotatebox{90}{
        \small $\quantile_{1-\alpha}\left( \ell^p_{Y,I}( \what{C}(X) ) \right) - \delta$}}
    
         \put(82, 24.5){
      \tikz{\path[draw=white, fill=white] (0, 0) rectangle (4cm, .5cm)}
    }
         \put(85, 23){
      \tikz{\path[draw=white, fill=white] (0, 0) rectangle (4cm, .4cm)}
    }
         \put(85, 21){
      \tikz{\path[draw=white, fill=white] (0, 0) rectangle (4cm, .4cm)}
    }
         \put(85, 19){
      \tikz{\path[draw=white, fill=white] (0, 0) rectangle (4cm, .4cm)}
    }
    
         \put(82, 24.5){
      \small \textbf{Method}
    }
         \put(85, 23){
      \scriptsize Alg.~\ref{alg:fixed-step-sequence-partial-loss}
    }
         \put(85, 21){
      \scriptsize Alg.~\ref{alg:nested-uniform-probe-based-conformalization}
    }
         \put(85, 19){
      \scriptsize Alg.~\ref{alg:step-up-probe-based-conformalization}
    }

         \put(82, 16.7){
      \tikz{\path[draw=white, fill=white] (0, 0) rectangle (4cm, .3cm)}
    }
         \put(85, 14.7){
      \tikz{\path[draw=white, fill=white] (0, 0) rectangle (4cm, .4cm)}
    }
         \put(85, 13.8){
      \tikz{\path[draw=white, fill=white] (0, 0) rectangle (4cm, .4cm)}
    }
    
             \put(82, 17.6){
      \small \textbf{Predictive set sequence}
    }
         \put(85, 15.8){
      \scriptsize $\{ C_{\eta^\star(X,\delta)}(X) \}_{0 \le \delta \le 1}$~\eqref{eqn:adaptive-threshold-choice}
    }
         \put(85, 13.6){
      \scriptsize $\{ C_\lambda^\score(X) \}_{\lambda \ge 0}$~\eqref{eqn:threshold-pred-set}
      }
  \end{overpic}
  \caption{Results for ImageNet dataset~\cite{DengDoSoLiLiFe09}. 
 Difference between the $1-\alpha$-quantile on the test set of the loss $\ell^p_{Y,I}(C(X))$~\eqref{eqn:user-probe-loss} and $\delta$,
  for different values of $\alpha$ and $\delta$ and respective methods.
  We average our results over $10$ runs, with error bars displaying the interquartile range.
  A predictive set mapping $C$ provides adequate control of the loss if that difference falls below the $y=0$ line, since we then have $\P(\ell^p_{Y,I}( C(X)) > \delta) \le \alpha$.
  }
  \label{fig:imagenet-quantile-loss}
\end{figure}

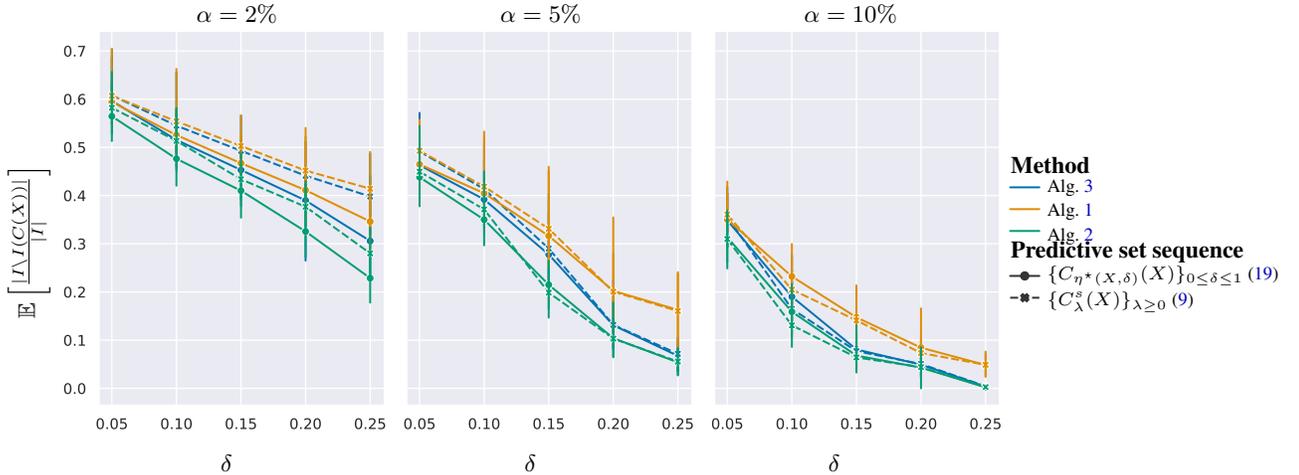
\begin{figure}
 \centering
  \begin{overpic}[
  				scale=0.5]{%
     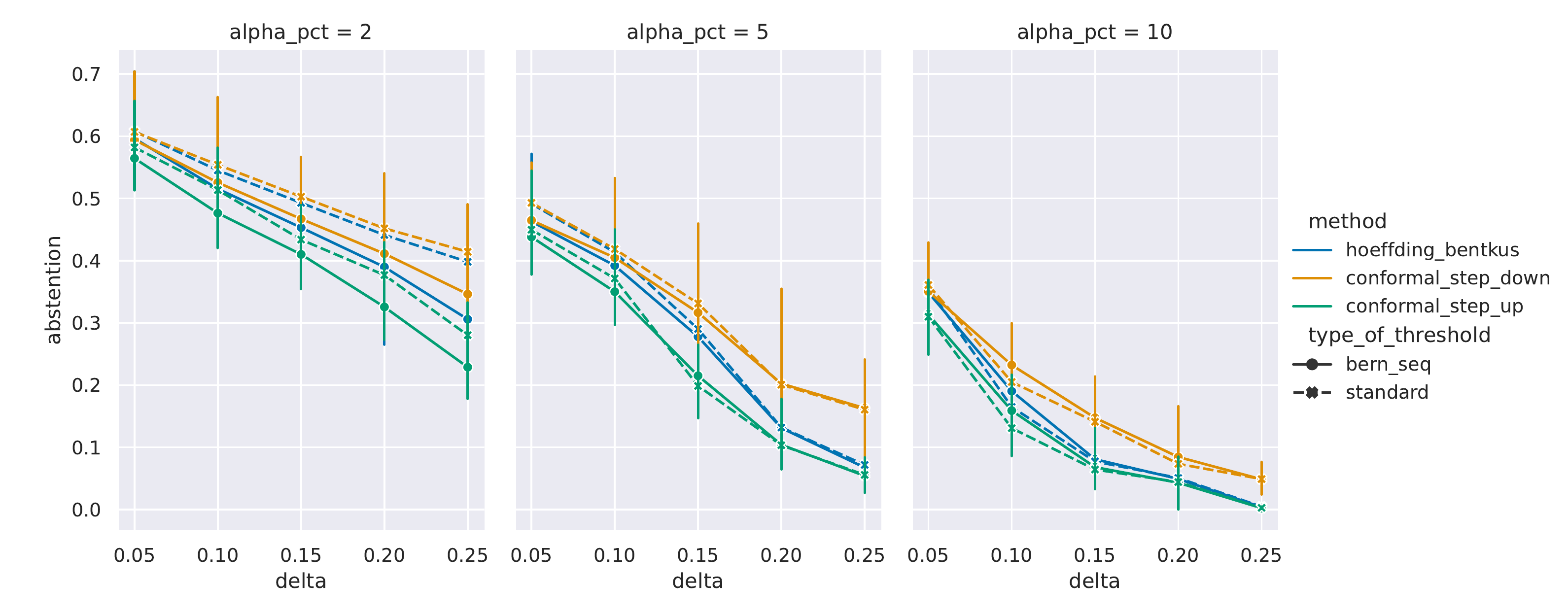}
     
    \put(15, 0){
      \tikz{\path[draw=white, fill=white] (0, 0) rectangle (4cm, .5cm)}
    }
    
        \put(40, 0){
      \tikz{\path[draw=white, fill=white] (0, 0) rectangle (4cm, .5cm)}
    }
    
        \put(62, 0){
      \tikz{\path[draw=white, fill=white] (0, 0) rectangle (4cm, .5cm)}
    }
    
    \put(17, 0){
       \small $\delta$
    }
        \put(42, 0){
       \small $\delta$
    }
        \put(67, 0){
       \small $\delta$
    }

        \put(10, 36.3){
      \tikz{\path[draw=white, fill=white] (0, 0) rectangle (4cm, .5cm)}
    }
    
        \put(32, 36.3){
      \tikz{\path[draw=white, fill=white] (0, 0) rectangle (4cm, .5cm)}
    }
    
        \put(60, 36.3){
      \tikz{\path[draw=white, fill=white] (0, 0) rectangle (4cm, .5cm)}
    }
    
    \put(15, 37){
       \small $\alpha = 2\%$
    }
        \put(40, 37){
       \small $\alpha = 5\%$
    }
        \put(65, 37){
       \small $\alpha = 10\%$
    }
    
    \put(70, 0){
      \tikz{\path[draw=white, fill=white] (0, 0) rectangle (4cm, .3cm)}
    }

    \put(0, 13){
      \tikz{\path[draw=white, fill=white] (0, 0) rectangle (.6cm, 5cm)}
    }
    \put(0, 12){\rotatebox{90}{
        \small $\E \left[ \frac{\left| I \setminus I(C(X)) \right|}{|I|} \right] $}}
    
         \put(82, 24.5){
      \tikz{\path[draw=white, fill=white] (0, 0) rectangle (4cm, .5cm)}
    }
         \put(85, 23){
      \tikz{\path[draw=white, fill=white] (0, 0) rectangle (4cm, .4cm)}
    }
         \put(85, 21){
      \tikz{\path[draw=white, fill=white] (0, 0) rectangle (4cm, .4cm)}
    }
         \put(85, 19){
      \tikz{\path[draw=white, fill=white] (0, 0) rectangle (4cm, .4cm)}
    }
    
         \put(82, 24.5){
      \small \textbf{Method}
    }
         \put(85, 23){
      \scriptsize Alg.~\ref{alg:fixed-step-sequence-partial-loss}
    }
         \put(85, 21){
      \scriptsize Alg.~\ref{alg:nested-uniform-probe-based-conformalization}
    }
         \put(85, 19){
      \scriptsize Alg.~\ref{alg:step-up-probe-based-conformalization}
    }

         \put(82, 16.7){
      \tikz{\path[draw=white, fill=white] (0, 0) rectangle (4cm, .3cm)}
    }
         \put(85, 14.7){
      \tikz{\path[draw=white, fill=white] (0, 0) rectangle (4cm, .4cm)}
    }
         \put(85, 13.8){
      \tikz{\path[draw=white, fill=white] (0, 0) rectangle (4cm, .4cm)}
    }
    
             \put(82, 17.6){
      \small \textbf{Predictive set sequence}
    }
         \put(85, 15.8){
      \scriptsize $\{ C_{\eta^\star(X,\delta)}(X) \}_{0 \le \delta \le 1}$~\eqref{eqn:adaptive-threshold-choice}
    }
         \put(85, 13.6){
      \scriptsize $\{ C_\lambda^\score(X) \}_{\lambda \ge 0}$~\eqref{eqn:threshold-pred-set}
    }     
          
  \end{overpic}
  \caption{Results for ImageNet dataset~\cite{DengDoSoLiLiFe09}. 
    Average abstention $1-\dfrac{|I \cap I(\what{C}_n(X)) | }{|I|}$ on the test set for different values of $\alpha$ and $\delta$ and respective methods.
  Lower abstention is better, since more informative.
  We average our results over $10$ runs, with error bars displaying the interquartile range.
  }
  \label{fig:imagenet-abstention-avg}
\end{figure}

\paragraph{Experimental results}
Figures~\ref{fig:imagenet-quantile-loss} and~\ref{fig:imagenet-loss-ecdf} show the $1-\alpha$-quantile of the loss and the empirical cumulative distribution of the loss on test examples. 
In accordance with our theorems,  they respectively confirm that  Alg.~\ref{alg:nested-uniform-probe-based-conformalization} and~\ref{alg:fixed-step-sequence-partial-loss} control the value-at-risk of the loss at level $\delta$, while Alg.~\ref{alg:step-up-probe-based-conformalization}, is as expected slightly too lax, especially when $\alpha$ is small.
These figures also outline the clear trade-off between loss and abstention: more conservative methods give less informative predictive sets.  

 In terms of predictive set sequences, Figures~\ref{fig:imagenet-abstention-avg} and~\ref{fig:imagenet-abstention-ecdf} indicate that adaptive thresholds are especially relevant for small values of $\alpha$, i.e.\ when one tries to provide stronger confidence guarantees, as they generally produce predictive sets with fewer abstention. 
A potential hypothesis here is that, when requiring the loss to be close to uniformly small (very small probability $\alpha \to 0$ of failed coverage),  it is especially beneficial to allow the instance-dependent thresholds: we can still provide valid predictive sets for very hard examples by choosing a larger threshold,  without hurting easier examples.

\bibliography{bib}

\begin{thebibliography}{18}
\providecommand{\natexlab}[1]{#1}
\providecommand{\url}[1]{\texttt{#1}}
\expandafter\ifx\csname urlstyle\endcsname\relax
  \providecommand{\doi}[1]{doi: #1}\else
  \providecommand{\doi}{doi: \begingroup \urlstyle{rm}\Url}\fi

\bibitem[Angelopoulos et~al.(2021)Angelopoulos, Bates, Cand\`{e}s, Jordan, and
  Lei]{AngelopoulosBaCaJoLe21}
A.~N. Angelopoulos, S.~Bates, E.~J. Cand\`{e}s, M.~I. Jordan, and L.~Lei.
\newblock Learn then test: Calibrating predictive algorithms to achieve risk
  control.
\newblock \emph{arXiv:2110.01052 [cs.LG]}, 2021.

\bibitem[Barber et~al.(2021)Barber, Cand\`{e}s, Ramdas, and
  Tibshirani]{BarberCaRaTi19a}
R.~F. Barber, E.~J. Cand\`{e}s, A.~Ramdas, and R.~J. Tibshirani.
\newblock The limits of distribution-free conditional predictive inference.
\newblock \emph{Information and Inference}, 10\penalty0 (2):\penalty0 455--482,
  2021.

\bibitem[Bates et~al.(2021)Bates, Angelopoulos, Lei, Malik, and
  Jordan]{BatesAnLeMaJo21}
S.~Bates, A.~Angelopoulos, L.~Lei, J.~Malik, and M.~I. Jordan.
\newblock Distribution-free, risk-controlling prediction sets.
\newblock \emph{Journal of the Association for Computing Machinery},
  68\penalty0 (6):\penalty0 43:1--43:34, 2021.

\bibitem[Bentkus(2004)]{Bentkus04}
V.~Bentkus.
\newblock On {H}oeffding's inequalities.
\newblock \emph{Annals of Probability}, 32\penalty0 (2), apr 2004.

\bibitem[Cao et~al.(2007)Cao, Qin, Liu, Tsai, and Li]{CaoQiLiTsLi07}
Z.~Cao, T.~Qin, T.-Y. Liu, M.-F. Tsai, and H.~Li.
\newblock Learning to rank: from pairwise approach to listwise approach.
\newblock In \emph{Proceedings of the 24th International Conference on Machine
  Learning}, pages 129--136, 2007.

\bibitem[Cauchois et~al.(2021)Cauchois, Gupta, and Duchi]{CauchoisGuDu21}
M.~Cauchois, S.~Gupta, and J.~Duchi.
\newblock Knowing what you know: valid and validated confidence sets in
  multiclass and multilabel prediction.
\newblock \emph{Journal of Machine Learning Research}, 22\penalty0
  (81):\penalty0 1--42, 2021.

\bibitem[Cauchois et~al.(2022)Cauchois, Gupta, Ali, and
  Duchi]{CauchoisGuAlDu22}
M.~Cauchois, S.~Gupta, A.~Ali, and J.~Duchi.
\newblock Predictive inference with weak supervision.
\newblock \emph{arXiv:2201.08315 [stat.ML]}, 2022.

\bibitem[Deng et~al.(2009)Deng, Dong, Socher, Li, Li, and
  Fei-Fei]{DengDoSoLiLiFe09}
J.~Deng, W.~Dong, R.~Socher, L.~Li, K.~Li, and L.~Fei-Fei.
\newblock Image{N}et: a large-scale hierarchical image database.
\newblock In \emph{Proceedings of the IEEE Conference on Computer Vision and
  Pattern Recognition}, pages 248--255, 2009.

\bibitem[Freund et~al.(2003)Freund, Iyer, Schapire, and Singer]{FreundIyScSi03}
Y.~Freund, R.~Iyer, R.~E. Schapire, and Y.~Singer.
\newblock Efficient boosting algorithms for combining preferences.
\newblock \emph{Journal of Machine Learning Research}, 4:\penalty0 933--969,
  2003.

\bibitem[Gupta et~al.(2022)Gupta, Kuchibhotla, and Ramdas]{GuptaKuRa22}
C.~Gupta, A.~K. Kuchibhotla, and A.~K. Ramdas.
\newblock Nested conformal prediction and quantile out-of-bag ensemble methods.
\newblock \emph{Pattern Recognition}, 127, 2022.
\newblock Special Issue on Conformal and Probabilistic Prediction with
  Applications.

\bibitem[Hoeffding(1963)]{Hoeffding63}
W.~Hoeffding.
\newblock Probability inequalities for sums of bounded random variables.
\newblock \emph{Journal of the American Statistical Association}, 58\penalty0
  (301):\penalty0 13--30, Mar. 1963.

\bibitem[Qin and Liu(2013)]{QiniLi13}
T.~Qin and T.~Liu.
\newblock Introducing {LETOR} 4.0 datasets.
\newblock \emph{arXiv:1306.2597 [cs.IR]}, 2013.

\bibitem[Recht et~al.(2019)Recht, Roelofs, Schmidt, and Shankar]{RechtRoScSh19}
B.~Recht, R.~Roelofs, L.~Schmidt, and V.~Shankar.
\newblock Do {I}mage{N}et classifiers generalize to {I}mage{N}et?
\newblock In \emph{Proceedings of the 36th International Conference on Machine
  Learning}, 2019.

\bibitem[Romano et~al.(2019)Romano, Patterson, and Cand\`{e}s]{RomanoPaCa19}
Y.~Romano, E.~Patterson, and E.~J. Cand\`{e}s.
\newblock Conformalized quantile regression.
\newblock In \emph{Advances in Neural Information Processing Systems 32}, 2019.

\bibitem[Romano et~al.(2020)Romano, Sesia, and Cand\`{e}s]{RomanoSeCa20}
Y.~Romano, M.~Sesia, and E.~J. Cand\`{e}s.
\newblock Classification with valid and adaptive coverage.
\newblock In \emph{Advances in Neural Information Processing Systems 33}, 2020.

\bibitem[Sadinle et~al.(2019)Sadinle, Lei, and Wasserman]{SadinleLeWa19}
M.~Sadinle, J.~Lei, and L.~Wasserman.
\newblock Least ambiguous set-valued classifiers with bounded error levels.
\newblock \emph{Journal of the American Statistical Association}, 114\penalty0
  (525):\penalty0 223--234, 2019.

\bibitem[Vovk(2012)]{Vovk12}
V.~Vovk.
\newblock Conditional validity of inductive conformal predictors.
\newblock In \emph{Proceedings of the Asian Conference on Machine Learning},
  volume~25 of \emph{Proceedings of Machine Learning Research}, pages 475--490,
  2012.

\bibitem[Vovk et~al.(2005)Vovk, Grammerman, and Shafer]{VovkGaSh05}
V.~Vovk, A.~Grammerman, and G.~Shafer.
\newblock \emph{Algorithmic Learning in a Random World}.
\newblock Springer, 2005.

\end{thebibliography}
\bibliographystyle{abbrvnat}

\newpage
\appendix
\section{Proofs of theorems}
\subsection{Proof of Theorem~\ref{thm:step-down-conform-validity}}
\label{subsec:proof-thm-step-down-conform-validity}
By definition of $\what{\lambda}_n$, we have $\P(S_{n+1} \le \what{\lambda}_n) \ge 1-\alpha$.
Now, by definition of $S_{n+1}$, for any $\lambda>S_{n+1}$, we have 
$\ell^p_{Y_{n+1}, I_{n+1}}(C^\score_{\lambda}(X_{n+1})) \le \delta$;  the function $\lambda \mapsto \ell^p_{Y_{n+1}, I_{n+1}} \left( C^\score_{\lambda}(X_{n+1}) \right)$ is right-continuous,  hence if $S_{n+1} \le \what{\lambda}_n$, we must have
\begin{align*}
\ell^p_{Y_{n+1}, I_{n+1}}(C^\score_{\what \lambda_n}(X_{n+1})) \le \delta,
\end{align*}
which yields the desired statement.

\subsection{Proof of Theorem~\ref{thm:step-up-conform-validity}}
\label{subsec:proof-thm-step-up-conform-validity}
To understand the amount of coverage loss,  notice there are only two ways in which we can have $\ell^p_{Y_{n+1}, I_{n+1}}(C^\score_{\what{\lambda}_n^\text{up}+\varepsilon}(X_{n+1})) > \delta$: either
\begin{align*}
\inf \{ \lambda > 0 \mid \ell^p_{Y_{n+1}, I_{n+1}}(C^\score_{\lambda }(X_{n+1})) \le \delta \} \eqdef \tilde S_{n+1} > \what{\lambda}_n^\text{up},
\end{align*} 
which occurs with probability at most $\alpha$, or $\tilde S_{n+1} \le \what{\lambda}_n^\text{up}$ and $C^\score_{\what{\lambda}_n^\text{up}+\varepsilon}(X_{n+1}) > \delta$, which means there exists $\lambda' =  \tilde S_{n+1} \le \what{\lambda}_n^\text{up}$ such that
\begin{align*}
\ell^p_{Y_{n+1}, I_{n+1}}(C^\score_{\lambda'}(X_{n+1})) \le \delta < \ell^p_{Y_{n+1}, I_{n+1}}(C^\score_{\what{\lambda}_n^\text{up}+\varepsilon}(X_{n+1})).
\end{align*}
Conditionally on the calibration set,  this event occurs with probability $\nmpcerr(\what{\lambda}_n^\text{up}, \varepsilon)$, 
which concludes the proof.

\subsection{Proof of Theorem~\ref{thm:fst-validity}}
\label{subsec:proof-thm-fst-validity}
First let $k^\star \defeq \max\left\{ k \in [N] \mid H_k \text{ holds} \right\}$.
If we have $\E \left[ \ell_{n+1} \left( \what{C}_n(X_{n+1}) \right) \mid  \what{C}_n \right]> \delta$,  then by definition of $k^\star$, we must have $\what{k}_n^\text{FST} \le k^\star$,
which in turns implies $p_{k^\star} \le \alpha_{\text{FST}}$ by definition of  $\what{k}_n^\text{FST}$.
Since $p_{k^\star}$ is a valid p-value for $H_{k^\star}$, this can only occur with probability less than $\alpha_{\text{FST}}$.

\section{Experimental set-up and additional figures}

\subsection{Experiment with LETOR~\cite{QiniLi13}}
\label{subsec:appendix-letor}
\begin{figure}[h]
 \centering
  \begin{overpic}[
  				scale=0.3]{%
     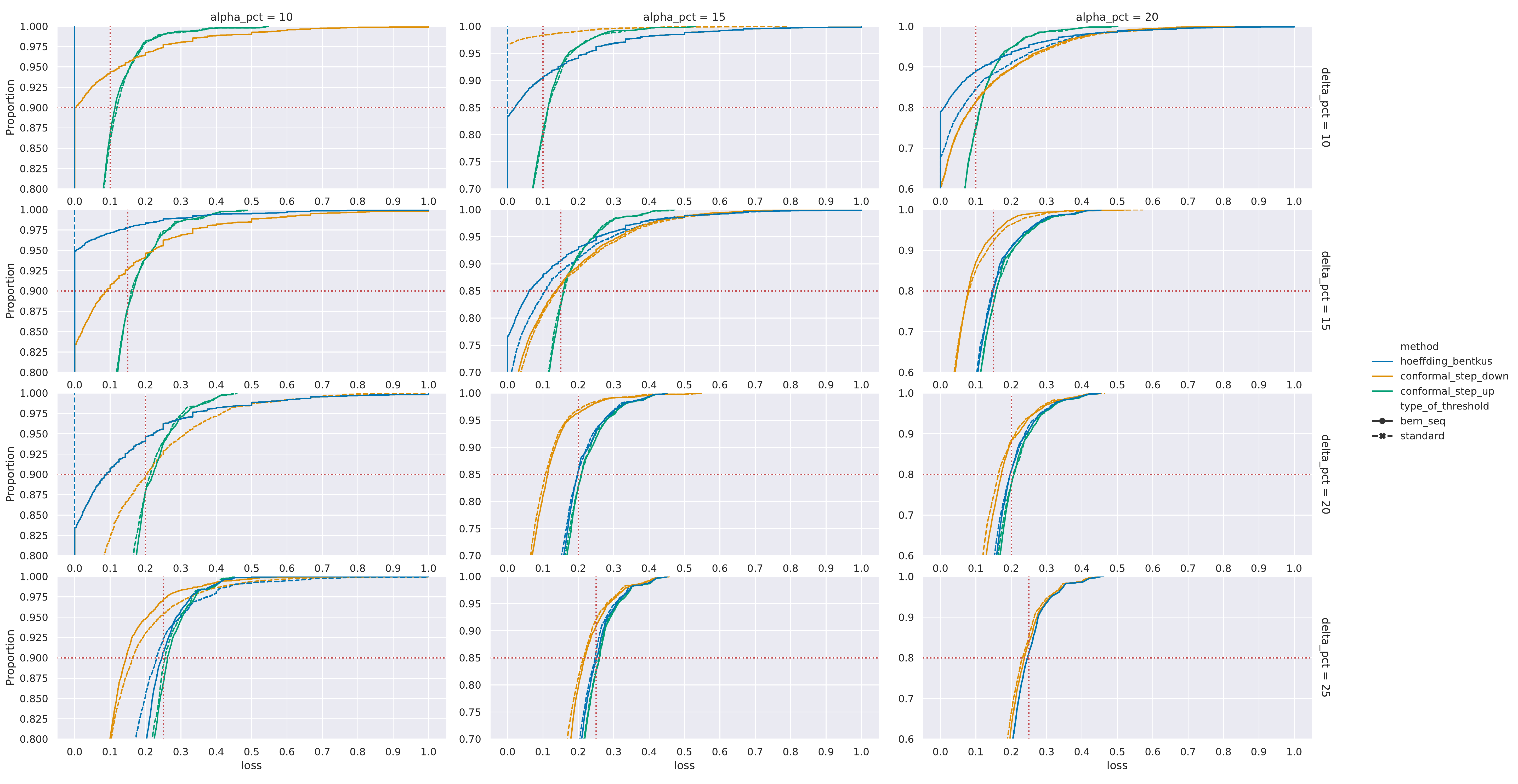}
 
     \put(14, 0){
      \tikz{\path[draw=white, fill=white] (0, 0) rectangle (4cm, .3cm)}
    }
    
        \put(40, 0){
      \tikz{\path[draw=white, fill=white] (0, 0) rectangle (4cm, .3cm)}
    }
    
        \put(62, 0){
      \tikz{\path[draw=white, fill=white] (0, 0) rectangle (4cm, .3cm)}
    }
    
    \put(15, 0){
       \small $t$
    }
        \put(44, 0){
       \small $t$
    }
        \put(73, 0){
       \small $t$
    }
    
    \put(-1, 0){
      \tikz{\path[draw=white, fill=white] (0, 0) rectangle (.3cm, 9cm)}
    }
    \put(-2, 20){\rotatebox{90}{
        \small $\P \left[\ell^p_{Y,I}(C(X))  \le t \right] $}}
        
     \put(86, 0){
      \tikz{\path[draw=white, fill=white] (0, 0) rectangle (.3cm, 9cm)}
    }
    \put(86.5, 48){\rotatebox{-90}{
        \small $\delta=10\%$}}
      
    \put(86.5, 36){\rotatebox{-90}{
        \small $\delta=15\%$}}
        
            \put(86.5, 24){\rotatebox{-90}{
        \small $\delta=20\%$}}
        
            \put(86.5, 12){\rotatebox{-90}{
        \small $\delta=25\%$}}

        \put(10, 50){
      \tikz{\path[draw=white, fill=white] (0, 0) rectangle (4cm, .5cm)}
    }
    
        \put(32, 50){
      \tikz{\path[draw=white, fill=white] (0, 0) rectangle (4cm, .5cm)}
    }
    
        \put(60, 50){
      \tikz{\path[draw=white, fill=white] (0, 0) rectangle (4cm, .5cm)}
    }
    
    \put(13, 51){
       \small $\alpha = 10\%$
    }
        \put(42, 51){
       \small $\alpha = 15\%$
    }
        \put(71, 51){
       \small $\alpha = 20\%$
    }

              \put(90, 28){
      \tikz{\path[draw=white, fill=white] (0, 0) rectangle (4cm, .3cm)}
    }
         \put(91.5, 27){
      \tikz{\path[draw=white, fill=white] (0, 0) rectangle (4cm, .2cm)}
    }
         \put(91.5, 26){
      \tikz{\path[draw=white, fill=white] (0, 0) rectangle (4cm, .2cm)}
    }
         \put(91.5, 25){
      \tikz{\path[draw=white, fill=white] (0, 0) rectangle (4cm, .2cm)}
    }
    
         \put(90, 28.5){
      \tiny \textbf{Method}
    }
         \put(91.5, 27.5){
      \tiny Alg.~\ref{alg:fixed-step-sequence-partial-loss}
    }
         \put(91.5, 26.5){
      \tiny Alg.~\ref{alg:nested-uniform-probe-based-conformalization}
    }
         \put(91.5, 25.5){
      \tiny Alg.~\ref{alg:step-up-probe-based-conformalization}
    }

         \put(90, 24.5){
      \tikz{\path[draw=white, fill=white] (0, 0) rectangle (4cm, .1cm)}
    }
         \put(91.5, 23){
      \tikz{\path[draw=white, fill=white] (0, 0) rectangle (4cm, .2cm)}
    }
         \put(91.5, 22){
      \tikz{\path[draw=white, fill=white] (0, 0) rectangle (4cm, .2cm)}
    }
    
             \put(90, 24.5){
      \tiny \textbf{Pred. set sequence}
    }
         \put(91.5, 23.7){
      \tiny $C_{\eta^\star(\cdot,\delta)}$~\eqref{eqn:adaptive-threshold-choice}
    }
         \put(91.5, 22.3){
      \tiny $C_\lambda^\score$~\eqref{eqn:threshold-pred-set}
      }
     
  \end{overpic}
  \caption{Results for LETOR ranking dataset~\cite{QiniLi13}. 
  Empirical cumulative distribution of the loss $\ell^p_{Y,I}(C(X))$~\eqref{eqn:user-probe-loss} on the test set, for our different methods.
  The red dotted horizontal line is the $y = 1-\alpha$, whereas the vertical one is the line $x=\delta$: a predictive set is valid if it falls above the intersection of these two lines.
  Each plot shows a different value of $\alpha \ge 0$ and $\delta \ge 0$, averaged over $10$ independent runs.
  }
  \label{fig:letor-loss-ecdf}
\end{figure}
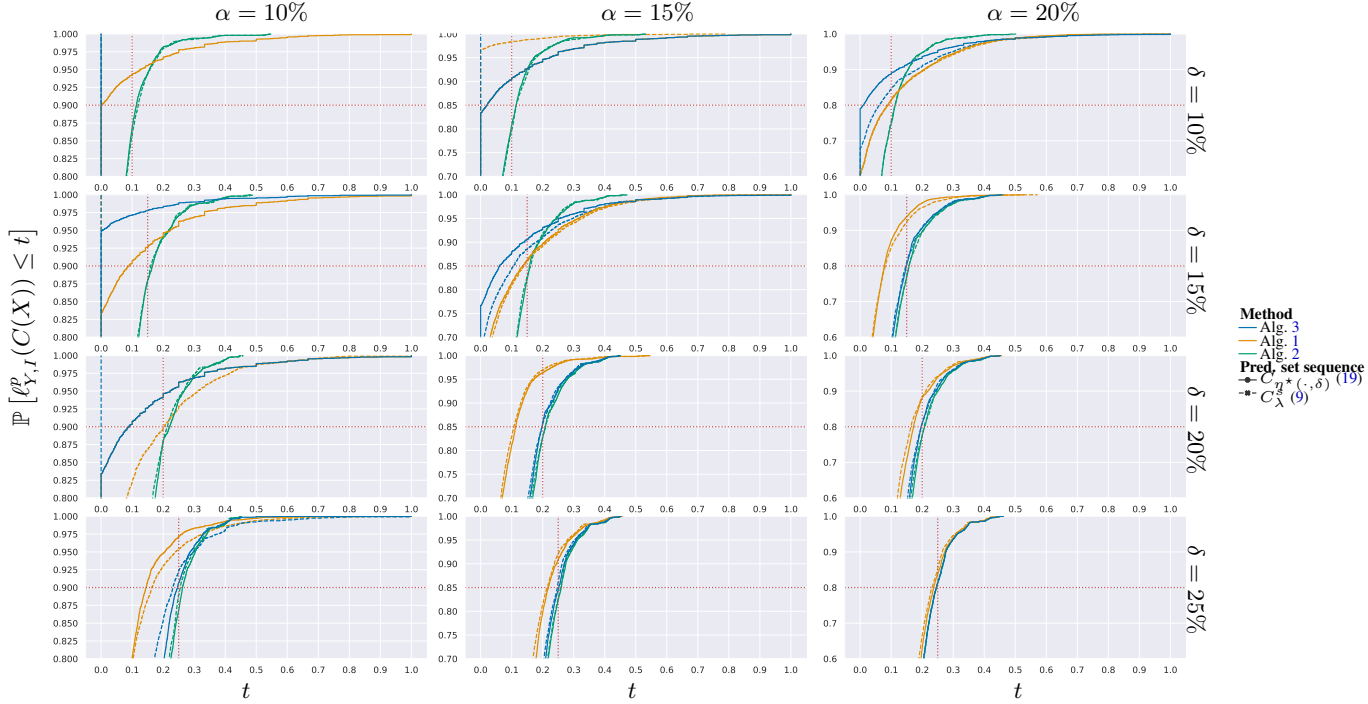

\paragraph{Learning relevance scores}
In these experiments, we learn individual relevance score functions $r(x,z) $ (with fully supervised training data) via the ListNet procedure~\cite{CaoQiLiTsLi07}, which we briefly describe here.
For a set $Z$ (with $K=|Z|$) of potential documents and respective relevance scores $\{ r_z \}_{z \in Z} \in \R^K$,  ListNet models the probability of a ranking $\pi \in \biject_K$ as    
\begin{align}
\label{eqn:listnet-model}
P_r(\pi) \defeq \prod_{y=1}^K \frac{\exp(r_{\pi(y)})}{\sum_{l=y}^k \exp(r_{\pi(l)})},
\end{align}
which gives each item $y \in \mc{Y}$ a top-1 probability (of ranking first) equal to
\begin{align*}
P^1_r(y)  \defeq P_r(\pi(1) = y) = \frac{\exp(r_{y})}{\sum_{l=1}^k \exp(r_{l})}.
\end{align*}
Given a training data set containing pairs $(X, R) \in \mc{X} \times \R^K$ of features/relevance scores,  we learn score mappings by minimizing the log-loss of the top-1 distribution over a set $\mc{F}$ of functions
\begin{align*}
\{ \hat{r}_y \}_{y \in [K]} \defeq \argmin_{ \tilde{r} \in \mc{F}^[K] } \left\{ \sum_{(X,R) \in \text{ training data}} \sum_{k=1}^K -P^1_{R}(k) \log\left( P^1_{\tilde{r}(X)}(k) \right)\right\}.
\end{align*}
Since in practice we only observe the ranking (or even a fraction of),  not the true per-item relevance scores, following common practice~\cite{CaoQiLiTsLi07, CauchoisGuAlDu22}, we use $R_y = K - ~ \text{the rank of the item} = K - Y^{-1}(y)$ as a proxy for our observed item-wise relevance scores when training our model.

\paragraph{Introducing weak supervision in the dataset}
The LETOR dataset contains (mostly) full rankings, so to evaluate our procedures, we sample a weakly supervised variant: for each $x$,  we sample binary comparisons (probe queries) as follows.
For each pair $(z_1, z_2)$ of potentially relevant document to $x$,  we independently add $(z_1,z_2)$  to the set $I$ of queries with a probability that only depends on the pair of relevance scores $(r(x, z_1),  r(x,z_2))$, while making sure to satisfy the following desiderata: we should only sample a small fraction of the overall $\choose{Z(x)}{2}$ possible pairs of documents (to remain in a weakly supervised setting) and should primarily sample pairs with high relevance scores, since a user is typically interested in relevant documents. 
Precisely,  we include the pair in the set of user queries with a probability equal to
\begin{align*}
1 - e^{-c_1 \left( r(x,z_1) \wedge r(x,z_2) \right) \left(1 + c_2 \left| r(x,z_1) - r(x, z_2) \right| \right)},
\end{align*} 
where $c_1 \defeq 0.05$ and $c_2 = 0.2$ are hyperparameters.
The term $ r(x,z_1) \wedge r(x,z_2) $ selects mostly pairs of relevant documents, while the second term $ \left| r(x,z_1) - r(x, z_2) \right| $ ensures that there is a significant relevance difference between the two documents.
We tuned the hyperparameters to sample around $2\%$ of all potential query comparisons, while ensuring that the average relevance difference between two documents reaches a certain level.

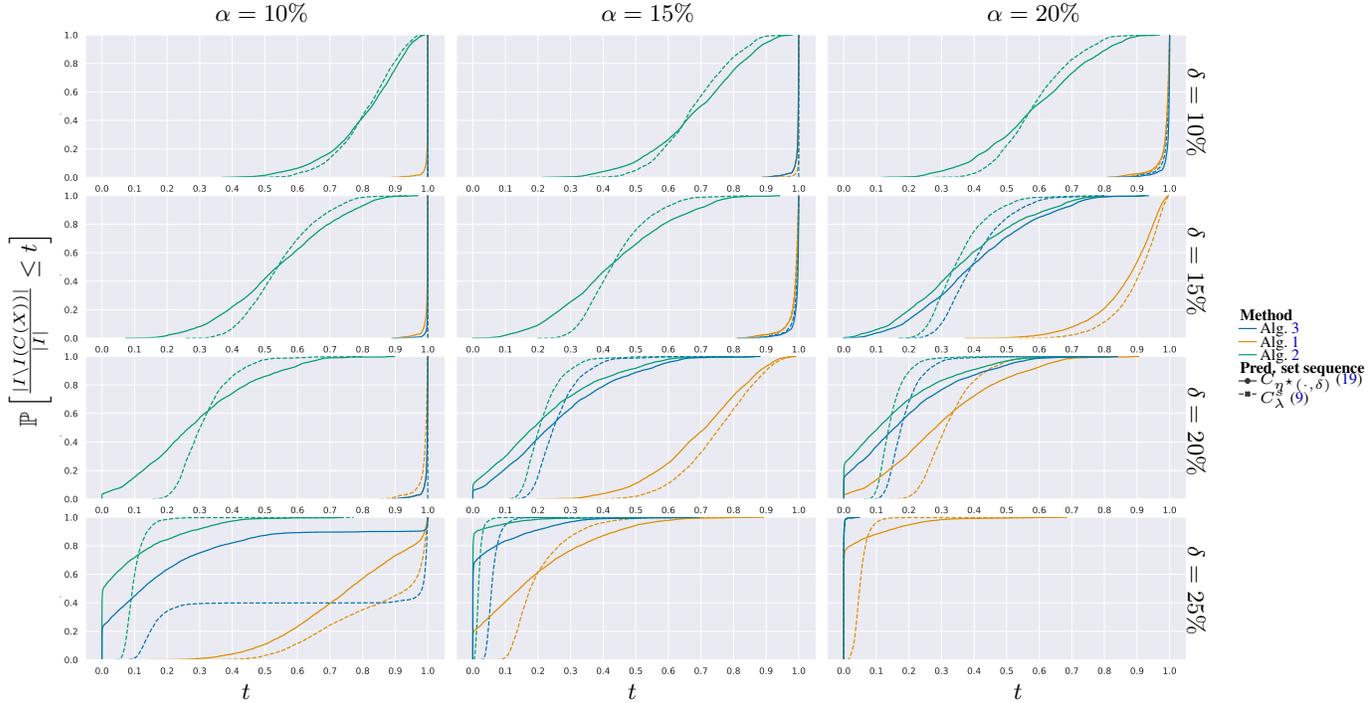
\begin{figure}
 \centering
  \begin{overpic}[
  				scale=0.3]{%
     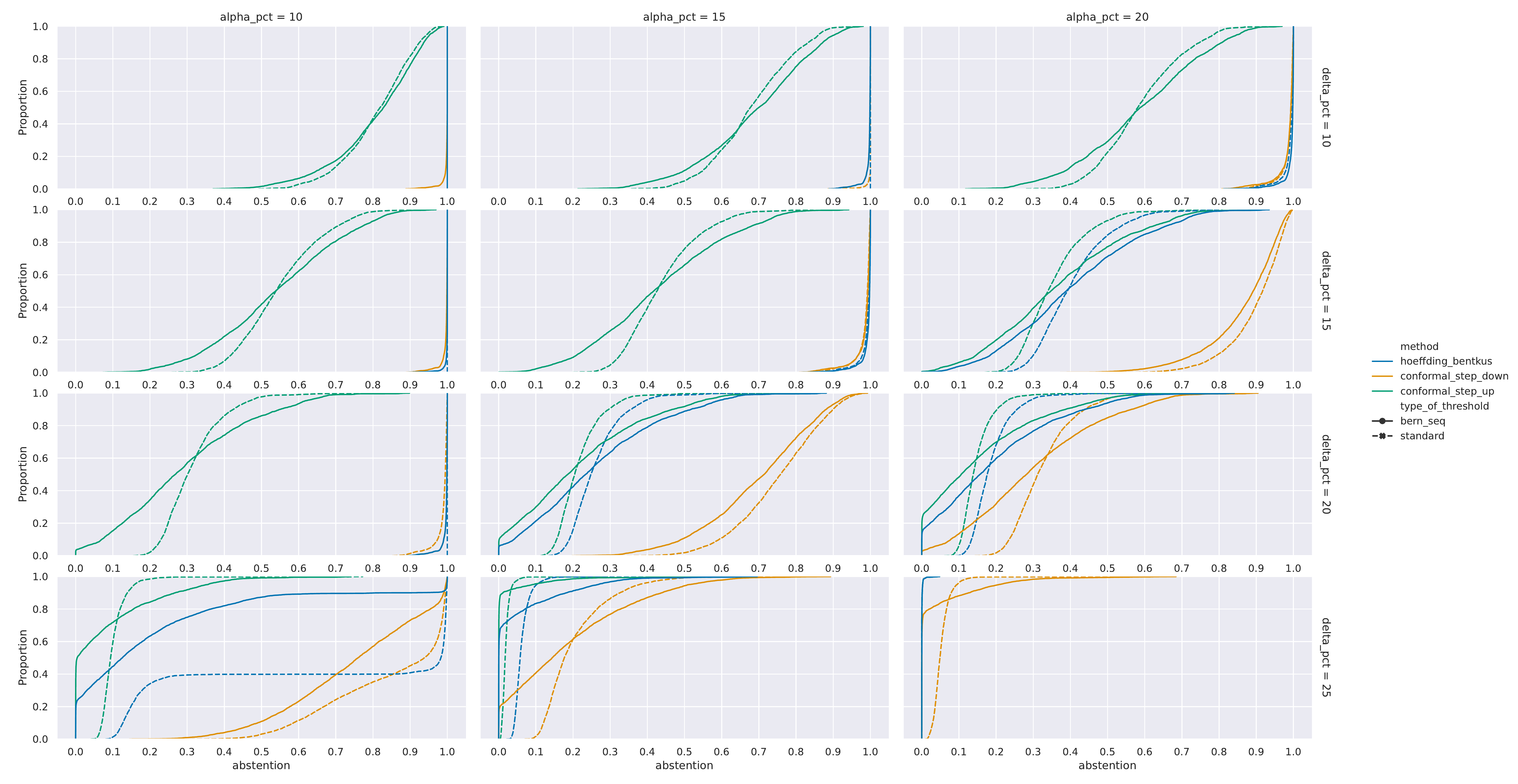}
      \put(14, 0){
      \tikz{\path[draw=white, fill=white] (0, 0) rectangle (4cm, .3cm)}
    }
    
        \put(40, 0){
      \tikz{\path[draw=white, fill=white] (0, 0) rectangle (4cm, .3cm)}
    }
    
        \put(62, 0){
      \tikz{\path[draw=white, fill=white] (0, 0) rectangle (4cm, .3cm)}
    }
    
    \put(15, 0){
       \small $t$
    }
        \put(44, 0){
       \small $t$
    }
        \put(73, 0){
       \small $t$
    }
    
    \put(-1, 0){
      \tikz{\path[draw=white, fill=white] (0, 0) rectangle (.4cm, 9cm)}
    }
    \put(-2, 20){\rotatebox{90}{
        \small $\P \left[ \frac{|I \setminus I(C(X))|}{|I|}  \le t \right] $}}
        
     \put(86, 0){
      \tikz{\path[draw=white, fill=white] (0, 0) rectangle (.3cm, 9cm)}
    }
    \put(86.5, 48){\rotatebox{-90}{
        \small $\delta=10\%$}}
      
    \put(86.5, 36){\rotatebox{-90}{
        \small $\delta=15\%$}}
        
            \put(86.5, 24){\rotatebox{-90}{
        \small $\delta=20\%$}}
        
            \put(86.5, 12){\rotatebox{-90}{
        \small $\delta=25\%$}}

        \put(10, 50){
      \tikz{\path[draw=white, fill=white] (0, 0) rectangle (4cm, .5cm)}
    }
    
        \put(32, 50){
      \tikz{\path[draw=white, fill=white] (0, 0) rectangle (4cm, .5cm)}
    }
    
        \put(60, 50){
      \tikz{\path[draw=white, fill=white] (0, 0) rectangle (4cm, .5cm)}
    }
    
    \put(13, 51){
       \small $\alpha = 10\%$
    }
        \put(42, 51){
       \small $\alpha = 15\%$
    }
        \put(71, 51){
       \small $\alpha = 20\%$
    }

              \put(90, 28){
      \tikz{\path[draw=white, fill=white] (0, 0) rectangle (4cm, .3cm)}
    }
         \put(91.5, 27){
      \tikz{\path[draw=white, fill=white] (0, 0) rectangle (4cm, .2cm)}
    }
         \put(91.5, 26){
      \tikz{\path[draw=white, fill=white] (0, 0) rectangle (4cm, .2cm)}
    }
         \put(91.5, 25){
      \tikz{\path[draw=white, fill=white] (0, 0) rectangle (4cm, .2cm)}
    }
    
         \put(90, 28.5){
      \tiny \textbf{Method}
    }
         \put(91.5, 27.5){
      \tiny Alg.~\ref{alg:fixed-step-sequence-partial-loss}
    }
         \put(91.5, 26.5){
      \tiny Alg.~\ref{alg:nested-uniform-probe-based-conformalization}
    }
         \put(91.5, 25.5){
      \tiny Alg.~\ref{alg:step-up-probe-based-conformalization}
    }

         \put(90, 24.5){
      \tikz{\path[draw=white, fill=white] (0, 0) rectangle (4cm, .1cm)}
    }
         \put(91.5, 23){
      \tikz{\path[draw=white, fill=white] (0, 0) rectangle (4cm, .2cm)}
    }
         \put(91.5, 22){
      \tikz{\path[draw=white, fill=white] (0, 0) rectangle (4cm, .2cm)}
    }
    
             \put(90, 24.5){
      \tiny \textbf{Pred. set sequence}
    }
         \put(91.5, 23.7){
      \tiny $C_{\eta^\star(\cdot,\delta)}$~\eqref{eqn:adaptive-threshold-choice}
    }
         \put(91.5, 22.3){
      \tiny $C_\lambda^\score$~\eqref{eqn:threshold-pred-set}
      }
      
  \end{overpic}
  \caption{Results for LETOR ranking dataset~\cite{QiniLi13}. 
  Empirical cumulative distribution of the query abstention $\frac{\left| I \setminus I(C(X)) \right|}{|I|}$~\eqref{eqn:user-probe-loss} on the test set, for our different methods.
  Each plot shows a different value of $\alpha \ge 0$ and $\delta \ge 0$, averaged over $10$ independent runs.
  Methods with a c.d.f.\ in the upper left corner have a stochastically lower abstention, hence yield informative predictive sets.
  }
  \label{fig:letor-abstention-ecdf}
\end{figure}

\paragraph{Additional experimental set-up and results}
Our experiment investigates the trade-off between type I and type II error in function of the desired level of protection $(\alpha, \delta)$,  as well as the differences in abstention between our two different predictive set sequences. 
Our theorems guarantee that our methods control the $1-\alpha$-VaR of the loss at level $\delta$, or close to that, but we expect more accurate and adaptive predictive sets to abstain less.
Recalling that $\P( \ell_{n+1}(X_{n+1}) > \delta) \le \alpha$, Figure~\ref{fig:letor-abstention-ecdf} confirms that higher values of $\alpha$ and $\delta$ lead to lower abstention. 
Choosing $\alpha$ and $\delta$ sufficiently low even results in full abstention: the model cannot yield predictions at this level of confidence.

The comparison between both predictive set sequences $\{ C^\score_\lambda \}_{\lambda \ge 0}$~\eqref{eqn:predictive-set-probe-adapted} and $\{ C_{\eta^\star(\cdot,  \delta)}(\cdot) \}_{0\le \delta \le 1}$ (Alg.~\ref{alg:bernoulli-pred-set}) reveals a few notable findings.  
The distribution of the loss between both sequences is often close, except when using Alg.~\ref{alg:nested-uniform-probe-based-conformalization} to conformalize, in which case the sequence we construct in Alg.~\ref{alg:bernoulli-pred-set} appears to produce less conservative sets.
The distribution of the abstention varies more considerably, perhaps not in the direction we initially thought of: it presents more spread and a higher range across all test examples when using adaptive thresholds than when using a constant threshold for every example. 
However, Figure~\ref{fig:letor-abstention-avg} suggests that the abstention decreases on average when using adaptive thresholds, which conforms more to our initial intuition.

\subsection{Experiment with ImageNet~\cite{DengDoSoLiLiFe09}}
\label{subsec:appendix-imagenet}

\paragraph{Experimental data}
The data consists of $5 \times 10^4$ validation examples from the initial ImageNet data set~\cite{DengDoSoLiLiFe09}, as well as an additional $10^4$ examples from the ImageNet V2 data set~\cite{RechtRoScSh19}, to allow for varying levels of image difficulty.
In each trial, we choose $5 \times 10^4$ instances to validate on from the pooled sample, and test on the $10^4$ remaining ones.
We use as predictor a pre-trained ResNet-50 model, where each node $v \in \mc{T}$ receives the sum of all leaf nodes probabilities in the its subtree as a probability score, e.g., $p_{\text{car}} = p_{\text{passenger car}} + p_{\text{freight car}}$. 
We use $s_v(x) \defeq \log \frac{p_v(x)}{1-p_v(x)}$ as a final predictive score for query $\varphi_v$, and $\what{\pi}_v(x) =  \min\left(p_v(x), 1-p_v(x) \right)$ as an estimate of the probability of error.

\paragraph{Weak supervision details}

\label{subsec:appendix-imagenet}
\begin{figure}
 \centering
  \begin{overpic}[
  				scale=0.5]{%
     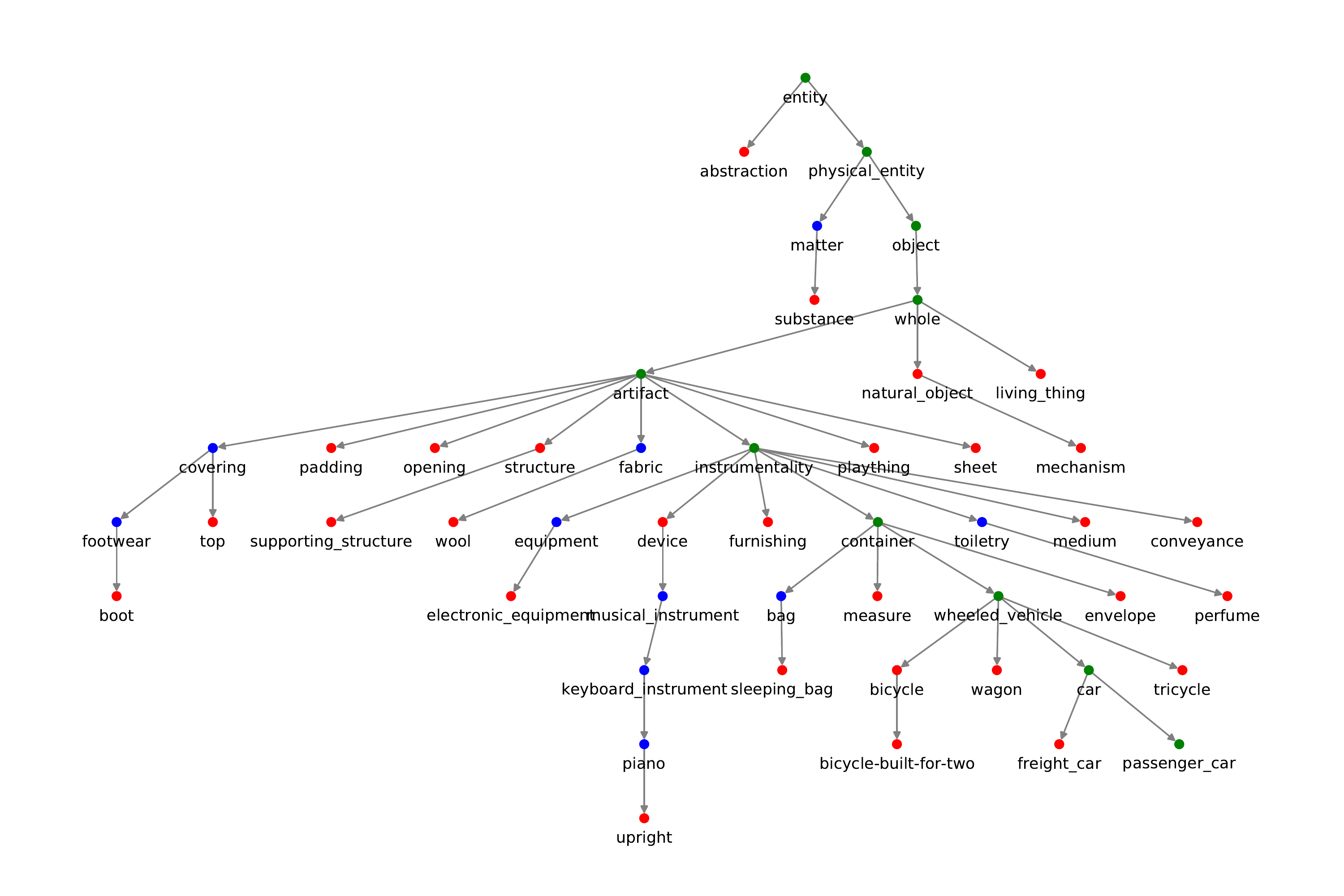}
  \end{overpic}
  \caption{Instance of a weakly supervised example for ImageNet dataset~\cite{DengDoSoLiLiFe09}. 
This particular image is a passenger car. 
Every node represents a potential user query of the form: "Is this node an ancestor of the true label?''.  Green and red nodes represent actual queries of interest for the user, whereas blue nodes are potential queries that this particular user did not select.
  }
  \label{fig:imagenet-query-instance}
\end{figure}

To every image then corresponds a set $I$ of queries, of which Figure~\ref{fig:imagenet-query-instance} provides a concrete example.  
For instance, if the image represents a tray,  the user may ask whether the image is a bag, or a receptacle, or more generally an artifact,  or physical entity. 
In this context,  our predictive set picks a subset of queries such that it can correctly answer a given fraction of these.

Our experimental set-up to select relevant queries for each image is as follows: for each image with label $y$, and each node $v \in \mc{V}$ in the tree, we select query $\varphi_v$ independently with a probability depending on the distances from $y$ and $v$ to their closest common ancestor $\text{ComAnc}(y,v)$.
The conceit here is that a user will likely ask queries about classes that bear some connection to the actual label, and so should be close in the tree.
With $\dist_\mc{T}$ the classical tree distance in $\mc{T}$, we select $v$ with probability
\begin{align*}
\min\left(1,  a \cdot e^{-b \dist_\mc{T}(y,\text{ComAnc}(y,v)) - c \dist_\mc{T}(v,\text{ComAnc}(y,v))} \right),
\end{align*}
where $a, b, c \ge 0$ are hyperparameters. 
We present results with $a=2$,  $b=.1$ and $c=1.5$, which produces $30$ queries on average per instance and samples nodes on the path to the true label more often than others; similar results holds with other hyperparameters.

\begin{figure}[h]
 \centering
  \begin{overpic}[
  				scale=0.3]{%
     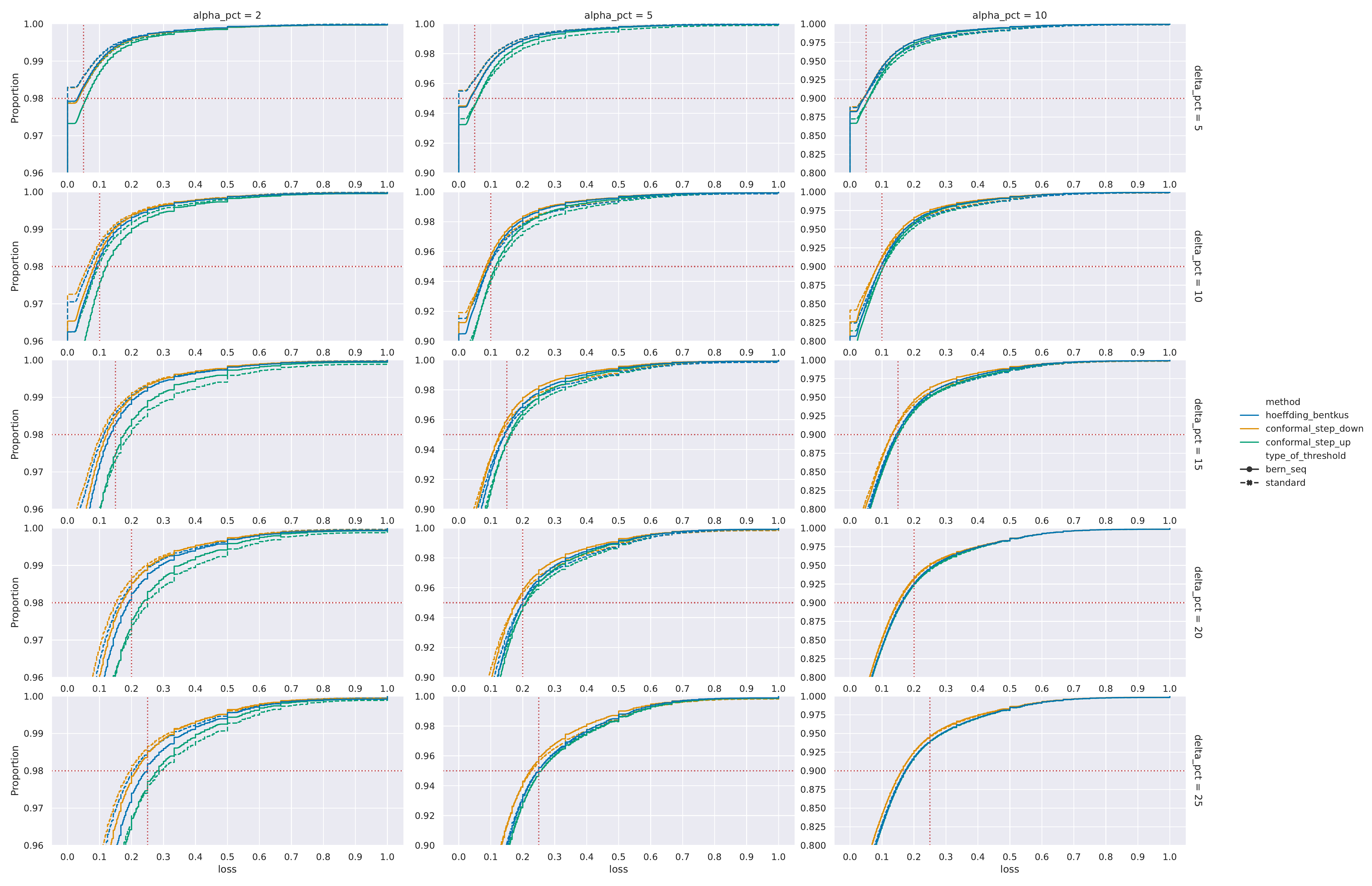}
      \put(14, 0){
      \tikz{\path[draw=white, fill=white] (0, 0) rectangle (4cm, .3cm)}
    }
    
        \put(40, 0){
      \tikz{\path[draw=white, fill=white] (0, 0) rectangle (4cm, .3cm)}
    }
    
        \put(62, 0){
      \tikz{\path[draw=white, fill=white] (0, 0) rectangle (4cm, .3cm)}
    }
    
    \put(15, 0){
       \small $t$
    }
        \put(44, 0){
       \small $t$
    }
        \put(73, 0){
       \small $t$
    }
    
    \put(-1, 0){
      \tikz{\path[draw=white, fill=white] (0, 0) rectangle (.35cm, 12cm)}
    }
    \put(-1, 25){\rotatebox{90}{
        \small $\P \left[\ell^p_{Y,I}(C(X))  \le t \right] $}}
        
     \put(86, 0){
      \tikz{\path[draw=white, fill=white] (0, 0) rectangle (.3cm, 12cm)}
    }
        \put(86.5, 60){\rotatebox{-90}{
        \small $\delta=5\%$}}
        
    \put(86.5, 48){\rotatebox{-90}{
        \small $\delta=10\%$}}
      
    \put(86.5, 36){\rotatebox{-90}{
        \small $\delta=15\%$}}
        
            \put(86.5, 24){\rotatebox{-90}{
        \small $\delta=20\%$}}
        
            \put(86.5, 12){\rotatebox{-90}{
        \small $\delta=25\%$}}

        \put(10, 63){
      \tikz{\path[draw=white, fill=white] (0, 0) rectangle (4cm, .5cm)}
    }
    
        \put(32, 63){
      \tikz{\path[draw=white, fill=white] (0, 0) rectangle (4cm, .5cm)}
    }
    
        \put(60, 63){
      \tikz{\path[draw=white, fill=white] (0, 0) rectangle (4cm, .5cm)}
    }
    
    \put(13, 63){
       \small $\alpha = 2\%$
    }
        \put(42, 63){
       \small $\alpha = 5\%$
    }
        \put(71, 63){
       \small $\alpha = 10\%$
    }

              \put(90, 34.5){
      \tikz{\path[draw=white, fill=white] (0, 0) rectangle (4cm, .3cm)}
    }
         \put(91.5, 33.3){
      \tikz{\path[draw=white, fill=white] (0, 0) rectangle (4cm, .2cm)}
    }
         \put(91.5, 32.3){
      \tikz{\path[draw=white, fill=white] (0, 0) rectangle (4cm, .2cm)}
    }
         \put(91.5, 31.3){
      \tikz{\path[draw=white, fill=white] (0, 0) rectangle (4cm, .2cm)}
    }
    
         \put(90, 34.8){
      \tiny \textbf{Method}
    }
         \put(91.5, 33.8){
      \tiny Alg.~\ref{alg:fixed-step-sequence-partial-loss}
    }
         \put(91.5, 32.8){
      \tiny Alg.~\ref{alg:nested-uniform-probe-based-conformalization}
    }
         \put(91.5, 31.8){
      \tiny Alg.~\ref{alg:step-up-probe-based-conformalization}
    }

         \put(90, 30.8){
      \tikz{\path[draw=white, fill=white] (0, 0) rectangle (4cm, .1cm)}
    }
         \put(91.5, 29.3){
      \tikz{\path[draw=white, fill=white] (0, 0) rectangle (4cm, .3cm)}
    }
         \put(91.5, 28.3){
      \tikz{\path[draw=white, fill=white] (0, 0) rectangle (4cm, .2cm)}
    }
    
             \put(90, 30.8){
      \tiny \textbf{Pred. set sequence}
    }
         \put(91.5, 30){
      \tiny $C_{\eta^\star(\cdot,\delta)}$~\eqref{eqn:adaptive-threshold-choice}
    }
         \put(91.5, 28.6){
      \tiny $C_\lambda^\score$~\eqref{eqn:threshold-pred-set}
      }
  \end{overpic}
  \caption{Results for ImageNet dataset~\cite{DengDoSoLiLiFe09}.   
    Empirical cumulative distribution of the loss $\ell^p_{Y,I}(C(X))$~\eqref{eqn:user-probe-loss} on the test set, for our different methods.
  The red dotted horizontal line is the $y = 1-\alpha$, whereas the vertical one is the line $x=\delta$: a predictive set is valid if it falls above the intersection of these two lines.
  Each plot shows a different value of $\alpha \ge 0$ and $\delta \ge 0$, averaged over $10$ independent runs.
  }
  \label{fig:imagenet-loss-ecdf}
\end{figure}

\begin{figure}
 \centering
  \begin{overpic}[
  				scale=0.30]{%
     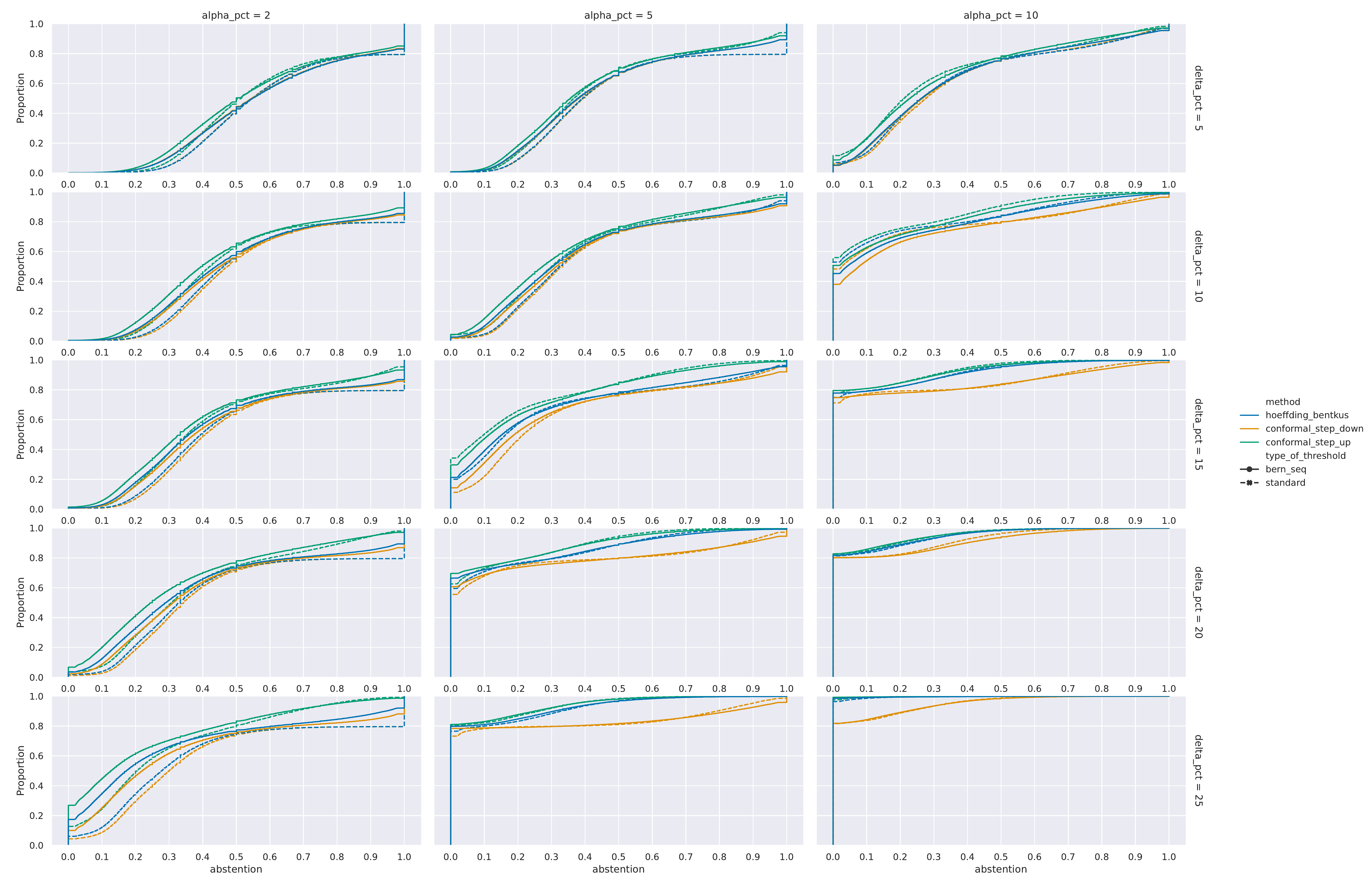}
        \put(14, 0){
      \tikz{\path[draw=white, fill=white] (0, 0) rectangle (4cm, .3cm)}
    }
    
        \put(40, 0){
      \tikz{\path[draw=white, fill=white] (0, 0) rectangle (4cm, .3cm)}
    }
    
        \put(62, 0){
      \tikz{\path[draw=white, fill=white] (0, 0) rectangle (4cm, .3cm)}
    }
    
    \put(15, 0){
       \small $t$
    }
        \put(44, 0){
       \small $t$
    }
        \put(73, 0){
       \small $t$
    }
    
    \put(-1, 0){
      \tikz{\path[draw=white, fill=white] (0, 0) rectangle (.4cm, 12cm)}
    }
    \put(-2, 25){\rotatebox{90}{
        \small $\P \left[ \frac{|I \setminus I(C(X))|}{|I|}  \le t \right] $}}
        
     \put(86, 0){
      \tikz{\path[draw=white, fill=white] (0, 0) rectangle (.3cm, 12cm)}
    }
        \put(86.5, 60){\rotatebox{-90}{
        \small $\delta=5\%$}}
        
    \put(86.5, 48){\rotatebox{-90}{
        \small $\delta=10\%$}}
      
    \put(86.5, 36){\rotatebox{-90}{
        \small $\delta=15\%$}}
        
            \put(86.5, 24){\rotatebox{-90}{
        \small $\delta=20\%$}}
        
            \put(86.5, 12){\rotatebox{-90}{
        \small $\delta=25\%$}}

        \put(10, 63){
      \tikz{\path[draw=white, fill=white] (0, 0) rectangle (4cm, .5cm)}
    }
    
        \put(32, 63){
      \tikz{\path[draw=white, fill=white] (0, 0) rectangle (4cm, .5cm)}
    }
    
        \put(60, 63){
      \tikz{\path[draw=white, fill=white] (0, 0) rectangle (4cm, .5cm)}
    }
    
    \put(13, 63){
       \small $\alpha = 2\%$
    }
        \put(42, 63){
       \small $\alpha = 5\%$
    }
        \put(71, 63){
       \small $\alpha = 10\%$
    }

              \put(90, 34.5){
      \tikz{\path[draw=white, fill=white] (0, 0) rectangle (4cm, .3cm)}
    }
         \put(91.5, 33.3){
      \tikz{\path[draw=white, fill=white] (0, 0) rectangle (4cm, .2cm)}
    }
         \put(91.5, 32.3){
      \tikz{\path[draw=white, fill=white] (0, 0) rectangle (4cm, .2cm)}
    }
         \put(91.5, 31.3){
      \tikz{\path[draw=white, fill=white] (0, 0) rectangle (4cm, .2cm)}
    }
    
         \put(90, 34.8){
      \tiny \textbf{Method}
    }
         \put(91.5, 33.8){
      \tiny Alg.~\ref{alg:fixed-step-sequence-partial-loss}
    }
         \put(91.5, 32.8){
      \tiny Alg.~\ref{alg:nested-uniform-probe-based-conformalization}
    }
         \put(91.5, 31.8){
      \tiny Alg.~\ref{alg:step-up-probe-based-conformalization}
    }

         \put(90, 30.8){
      \tikz{\path[draw=white, fill=white] (0, 0) rectangle (4cm, .1cm)}
    }
         \put(91.5, 29.3){
      \tikz{\path[draw=white, fill=white] (0, 0) rectangle (4cm, .3cm)}
    }
         \put(91.5, 28.3){
      \tikz{\path[draw=white, fill=white] (0, 0) rectangle (4cm, .2cm)}
    }
    
             \put(90, 30.8){
      \tiny \textbf{Pred. set sequence}
    }
         \put(91.5, 30){
      \tiny $C_{\eta^\star(\cdot,\delta)}$~\eqref{eqn:adaptive-threshold-choice}
    }
         \put(91.5, 28.6){
      \tiny $C_\lambda^\score$~\eqref{eqn:threshold-pred-set}
      }
  \end{overpic}
  \caption{Results for ImageNet dataset~\cite{DengDoSoLiLiFe09}. 
  Empirical cumulative distribution of the query abstention $\frac{\left| I \setminus I(C(X)) \right|}{|I|}$~\eqref{eqn:user-probe-loss} on the test set, for our different methods.
  Each plot shows a different value of $\alpha \ge 0$ and $\delta \ge 0$, averaged over $10$ independent runs.
  }
  \label{fig:imagenet-abstention-ecdf}
\end{figure}
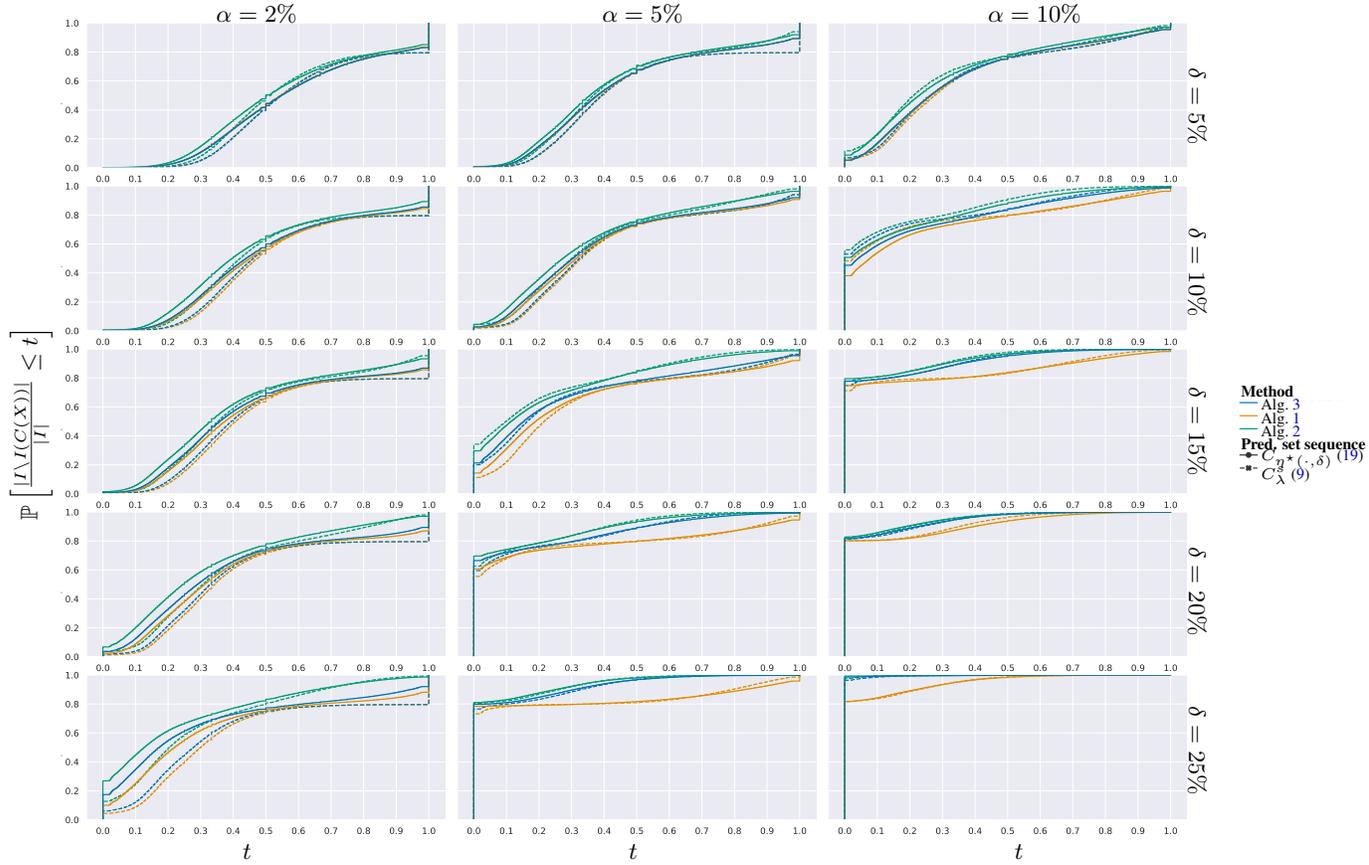
\end{document}